\pdfoutput=1
\documentclass[10pt,twocolumn,letterpaper]{article}

\usepackage{cvpr}
\usepackage{times}
\usepackage{epsfig}
\usepackage{graphbox} 

\usepackage{amsmath}
\usepackage{amssymb}
\usepackage{subfigure}
\usepackage[square,numbers]{natbib}
\usepackage{algorithm2e,algorithmic}

\usepackage[pagebackref=true,breaklinks=true,letterpaper=true,colorlinks,bookmarks=false]{hyperref}

\cvprfinalcopy 

\def\cvprPaperID{1362} 

\ifcvprfinal\fi
\begin{document}

\title{Synthesizing Normalized Faces from Facial Identity Features}

\author{Forrester Cole$^1$ David Belanger$^{1,2}$ Dilip Krishnan$^1$ Aaron Sarna$^1$ Inbar Mosseri$^1$ William T. Freeman$^{1,3}$ \\
$^1$Google, Inc. $^2$University of Massachusetts Amherst $^3$MIT CSAIL \\
\tt\small \{fcole, dbelanger, dilipkay, sarna, inbarm, wfreeman\}@google.com}

\maketitle

\begin{abstract}
   We present a method for synthesizing a frontal, neutral-expression image of a person's face given an input face photograph. This is achieved by learning to generate facial landmarks and textures from features extracted from a facial-recognition network. Unlike previous generative approaches, our encoding feature vector is largely invariant to lighting, pose, and facial expression. Exploiting this invariance, we train our decoder network using only frontal, neutral-expression photographs. Since these photographs are well aligned, we can decompose them into a sparse set of landmark points and aligned texture maps. The decoder then predicts landmarks and textures independently and combines them using a differentiable image warping operation. The resulting images can be used for a number of applications, such as analyzing facial attributes, exposure and white balance adjustment, or creating a 3-D avatar. 
\end{abstract}

\section{Introduction}
\label{sec:intro}
\begin{figure}[ht]
\vspace{-0.4cm}
\begin{tabular}{ccc}
  \includegraphics[width=0.3\columnwidth]{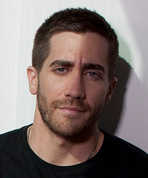} &
  \includegraphics[width=0.3\columnwidth]{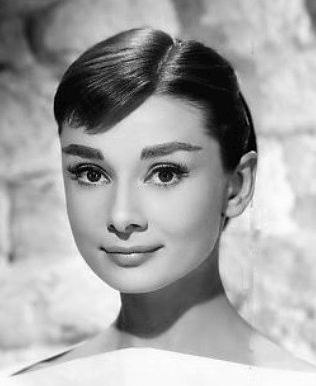} &
  \includegraphics[width=0.3\columnwidth]{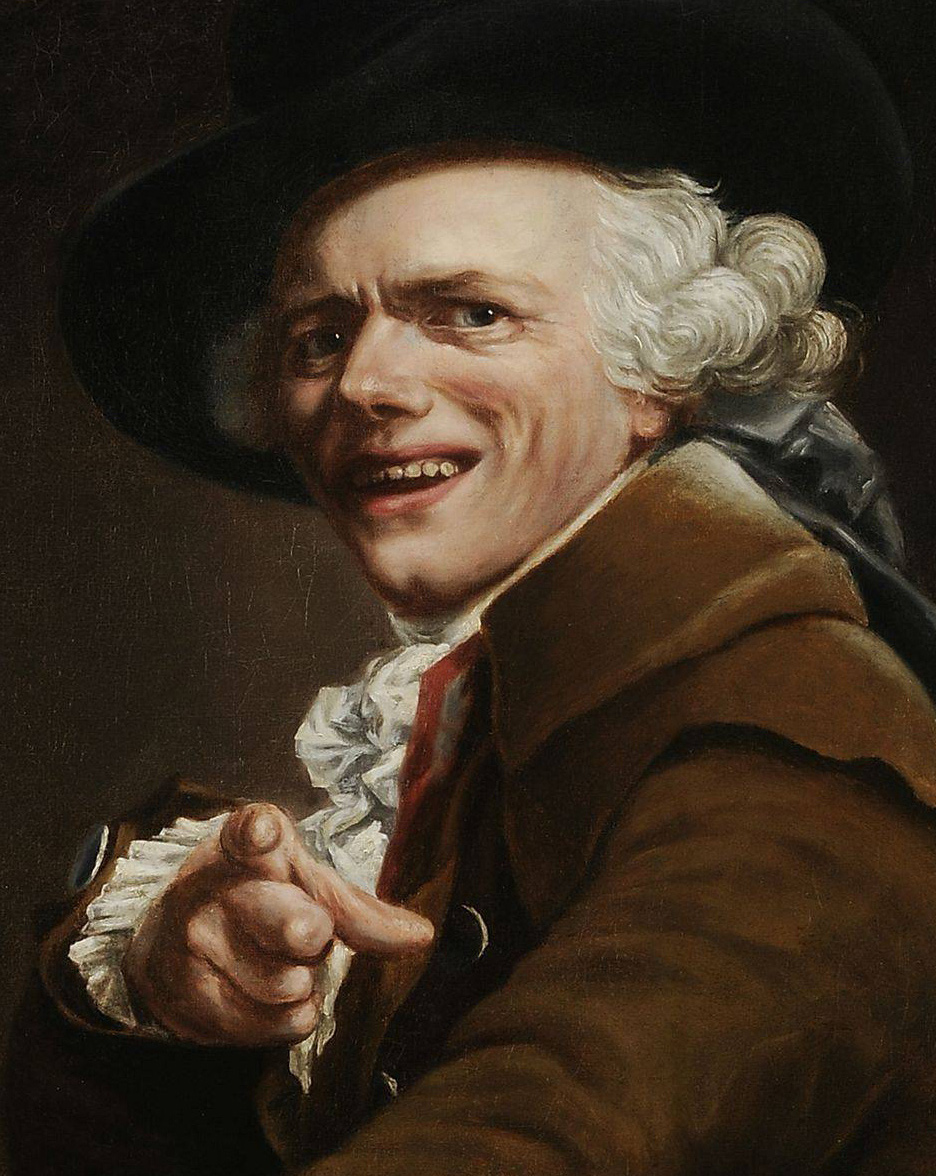} \\
  $\Downarrow$ & $\Downarrow$ & $\Downarrow$ \\
  1024-D features & 1024-D features & 1024-D features \\
  $\Downarrow$ & $\Downarrow$ & $\Downarrow$ \\
  \includegraphics[width=0.3\columnwidth]{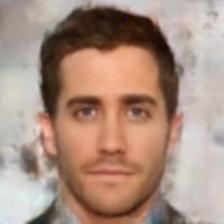} &
  \includegraphics[width=0.3\columnwidth]{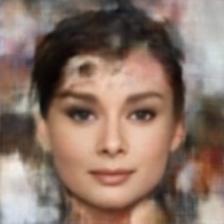} &
  \includegraphics[width=0.3\columnwidth]{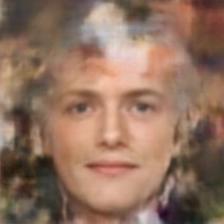} \\
  \includegraphics[width=0.3\columnwidth]{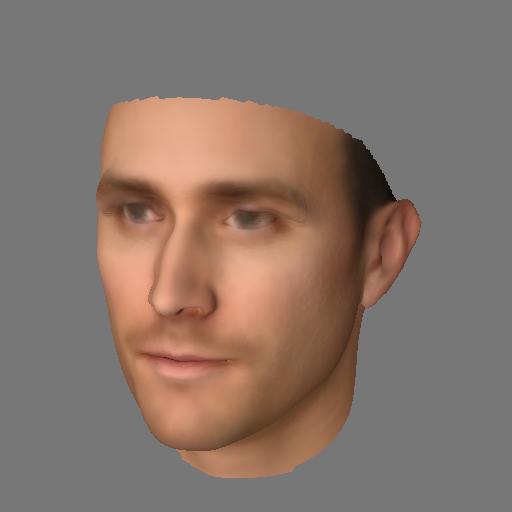} &
  \includegraphics[width=0.3\columnwidth]{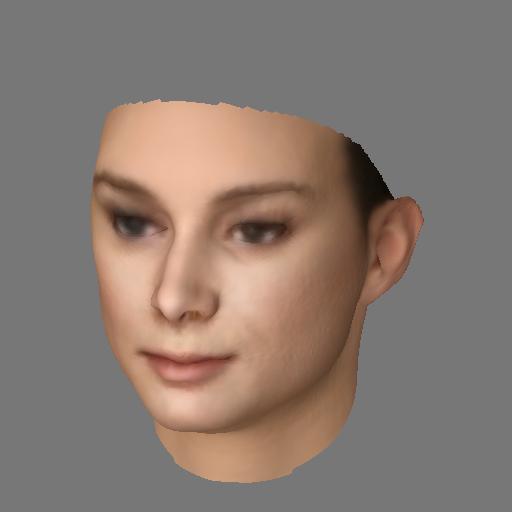} &
  \includegraphics[width=0.3\columnwidth]{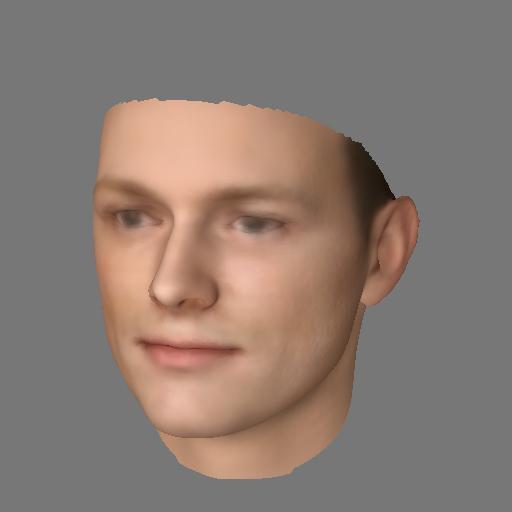} \\
\end{tabular}
\caption{Input photos (top) are encoded using a face recognition network~\cite{schroff2015facenet} into 1024-D feature vectors, then decoded into an image of the face using our decoder network (middle). 
The invariance of the encoder network to pose, lighting, and expression allows the decoder to produce a normalized face image. 
The resulting images can be easily fit to a 3-D model~\cite{blanz1999morphable} (bottom). 
Our method can even produce plausible reconstructions from black-and-white photographs and paintings of faces.} \label{fig:intro-examples}
\end{figure}

Recent work in computer vision has produced deep neural networks that are extremely effective at face recognition, achieving high accuracy over millions of identities~\cite{kemelmacher2015megaface}. These networks embed an input photograph in a high-dimensional feature space, where photos of the same person map to nearby points. The feature vectors produced by a network such as FaceNet~\cite{schroff2015facenet} are remarkably consistent across changes in pose, lighting, and expression. As is common with neural networks, however, the features are opaque to human interpretation. There is no obvious way to reverse the embedding and produce an image of a face from a given feature vector.

We present a method for mapping from facial identity features back to images of faces. This problem is hugely underconstrained: the output image has $150\times$ more dimensions than a FaceNet feature vector. Our key idea is to exploit the invariance of the facial identity features to pose, lighting, and expression by posing the problem as mapping from a feature vector to an evenly-lit, front-facing, neutral-expression face, which we call a \textit{normalized} face image. Intuitively, the mapping from identity to normalized face image is nearly one-to-one, so we can train a decoder network to learn it  (Fig.~\ref{fig:intro-examples}). We train the decoder network on carefully-constructed pairs of features and normalized face images. Our best results use FaceNet features, but the method produces similar results from features generated by the publicly-available VGG-Face network~\cite{parkhi2015deep}.

Because the facial identity features are so reliable, the trained decoder network is robust to a broad range of nuisance factors such as occlusion, lighting, and pose variation, and can even successfully operate on monochrome photographs or paintings. The robustness of the network sets it apart from related methods that directly \textit{frontalize} the face by warping the input image to a frontal pose~\cite{hassner2015effective, taigman2014deepface}, which cannot compensate for occlusion or lighting variation.

The consistency of the resulting normalized face allows a range of applications. For example, the neutral expression of the synthesized face and the facial landmark locations make it easy to fit a 3-D morphable model~\cite{blanz1999morphable} to create a virtual reality avatar  (Sec.~\ref{sec:3d_model_fitting}). Automatic color correction and white balancing can also be achieved by transforming the color of the input photograph to match the color of the predicted face (Sec.~\ref{sec:whitebalance}). Finally, our method can be used as an exploratory tool for visualizing what features are reliably captured by a facial recognition system.

Similar to the~\textit{active shape model} of~\citet{lanitis1995unified}, our decoder network explicitly decouples the face's geometry from its texture. In our case, the decoder produces both a registered texture image and the positions of facial landmarks as intermediate activations. Based on the landmarks, the texture is warped to obtain the final image. 

In developing our model, we tackle a few technical challenges.
First, end-to-end learning requires that the warping operation is \textit{differentiable}. We employ an efficient, easy-to-implement method based on spline interpolation. This allows us to compute FaceNet similarity between the input and output images as a training objective, which helps to retain perceptually-relevant details.

Second, it is difficult to obtain large amounts of front-facing, neutral-expression training data. In response, we employ a data-augmentation scheme that exploits the texture-shape decomposition, where we randomly morph the training images by interpolating with nearest neighbors. The augmented training set allows for fitting a high-quality neural network model using only 1K unique input images.  

The techniques introduced in this work, such as decomposition into geometry and texture, data augmentation, and differentiable warping, are applicable to domains other than face normalization.

\section{Background and Related Work}
\label{sec:background}

\subsection{Inverting Deep Neural Network Features}

The interest in understanding deep networks' predictions has led to several approaches for creating an image from a particular feature vector. One approach directly optimizes the image pixels by gradient descent~\cite{erhan2009visualizing,simonyan2013inside,mahendran2014understanding,yosinski2015understanding}, producing images similar to ``DeepDream''~\citep{deepdream}. Because the pixel space is so large relative to the feature space, optimization requires heavy regularization terms, such as total variation~\cite{mahendran2014understanding} or Gaussian blur~\cite{yosinski2015understanding}. The resulting images are intriguing, but not realistic. 

A second, more closely-related approach trains a feed-forward network to reverse a given embedding~\cite{zeiler2013visualizing, dosovitskiy2015inverting}. \citet{dosovitskiy2015inverting} pose this problem as constructing the most likely image given a feature vector. Our method, in contrast, uses the more restrictive criterion that the image must be a normalized face.

Perhaps the most relevant prior work is \citet{zhmoginov2016inverting}, which employs both iterative and and feed-forward methods for inverting FaceNet embeddings to recover an image of a face. While they require no training data, our method produces better fine-grained details.

\subsection{Active Appearance Model for Faces}

\begin{figure}
\begin{tabular}{ccc}
  \includegraphics[width=0.3\columnwidth]{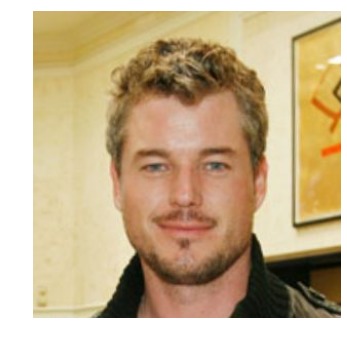} &
  \includegraphics[width=0.3\columnwidth]{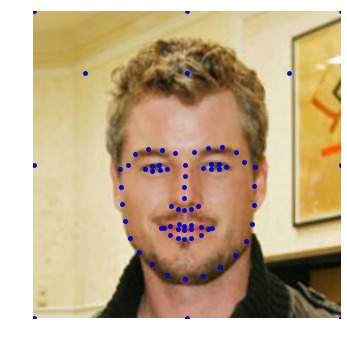} &
  \includegraphics[width=0.3\columnwidth]{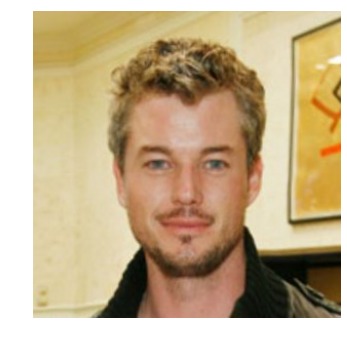} \\
\end{tabular}
\caption{From left to right: Input training image, detected facial landmark points, and the result of warping the input image to the mean face geometry.} \label{fig:warp-examples}
\end{figure}

The active appearance model of~\citet{cootes1998aam} and its extension to 3-D by~\citet{blanz1999morphable} provide parametric models for manipulating and generating face images. The model is fit to limited data by decoupling faces into two components: texture $T$ and the facial landmark geometry $L$. In Fig. ~\ref{fig:warp-examples} (middle), a set $L$ of landmark points (e.g., tip of nose) are detected. In Fig. ~\ref{fig:warp-examples} (right), the image is warped such that its landmarks are located at the training dataset's mean landmark locations $\bar{L}$. The warping operation aligns the textures so that, for example, the left pupil in every training image lies at the same pixel coordinates. 

In~\cite{cootes1998aam, blanz1999morphable}, the authors fit separate principal components analysis (PCA) models to the textures and geometry. These can be fit reliably using substantially less data than a PCA model on the raw images.  An individual face is described by the coefficients of the principal components of the landmarks and textures. To reconstruct the face, the coefficients are un-projected to obtain reconstructed landmarks and texture, then the texture is warped to the landmarks. 

There are various techniques for warping. For example, ~\citet{blanz1999morphable} define triangulations for both $L$ and $\bar{L}$ and apply an affine transformation for each triangle in $L$ to map it to the corresponding triangle in $\bar{L}$. In Sec.~\ref{sec:differentiable warping} we employ an alternative based on spline interpolation.



\subsection{FaceNet}
FaceNet~\cite{schroff2015facenet} maps from face images taken in the wild to 128-dimensional features. Its architecture is similar to the popular Inception model~\cite{szegedy2015going}. FaceNet is trained with a triplet loss: the embeddings of two pictures of person A should be more similar than the embedding of a picture of person A and a picture of person B. This loss encourages the model to capture aspects of a face pertaining to its identity, such geometry, and ignore factors of variation specific to the instant the image was captured, such as lighting, expression, pose, etc. FaceNet is trained on a very large dataset that encodes information about a wide variety of human faces.
Recently, models trained on publicly available data have approached or exceeded FaceNet's performance~\cite{parkhi2015deep}. Our method is agnostic to the source of the input features and produces similar results from features of the VGG-Face network as from FaceNet (Fig.~\ref{fig:frontalization}). 

We employ FaceNet both as a source of pretrained input features and as a source of a training loss: the input image and the generated image should have similar FaceNet embeddings. Loss functions defined via pretrained networks may be more correlated with perceptual, rather than pixel-level, differences ~\cite{DBLP:journals/corr/DosovitskiyB16,Johnson2016Perceptual}. 

\subsection{Face Frontalization}
Prior work in \textit{face frontalization} adopts a non-parametric approach to registering and normalizing face images taken in the wild~\cite{asthana2011pose,asthana2011fully,Yi_2013_CVPR,yi2013towards,taigman2014deepface,hassner2015effective}. Landmarks are detected on the input image and these are aligned to points on a reference 3-D or 2-D model. Then, the image is pasted on the reference model using non-linear warping. Finally, the rendered front-facing image can be fed to downstream models that were trained on front-facing images. The approach is largely parameter-free and does not require labeled training data, but does not normalize variation due to lighting, expression, or occlusion (Fig.~\ref{fig:frontalization}).

\subsection{Face Generation using Neural Networks}

Unsupervised learning of generative image models is an active research area, and many papers evaluate on the celebA dataset~\citep{liu2015faceattributes} of face images~\citep{liu2015faceattributes,larsen2015autoencoding,zhao2016energy,dinh2016density}. In these, the generated images are smaller and generally lower-quality than ours. Contrasting these approaches vs. our system is also challenging because they draw independent samples, whereas we generate images conditional on an input image. Therefore, we can not achieve high quality simply by memorizing certain prototypes.

\section{Autoencoder Model}
\label{sec:model}
We assume a training set of front-facing, neutral-expression training images. As preprocessing, we decompose each image into a texture $T$ and a set of landmarks $L$ using  off-the-shelf landmark detection tools and the warping technique of Sec.~\ref{sec:differentiable warping}.

\begin{figure}[!ht]
    \includegraphics[width=\columnwidth]{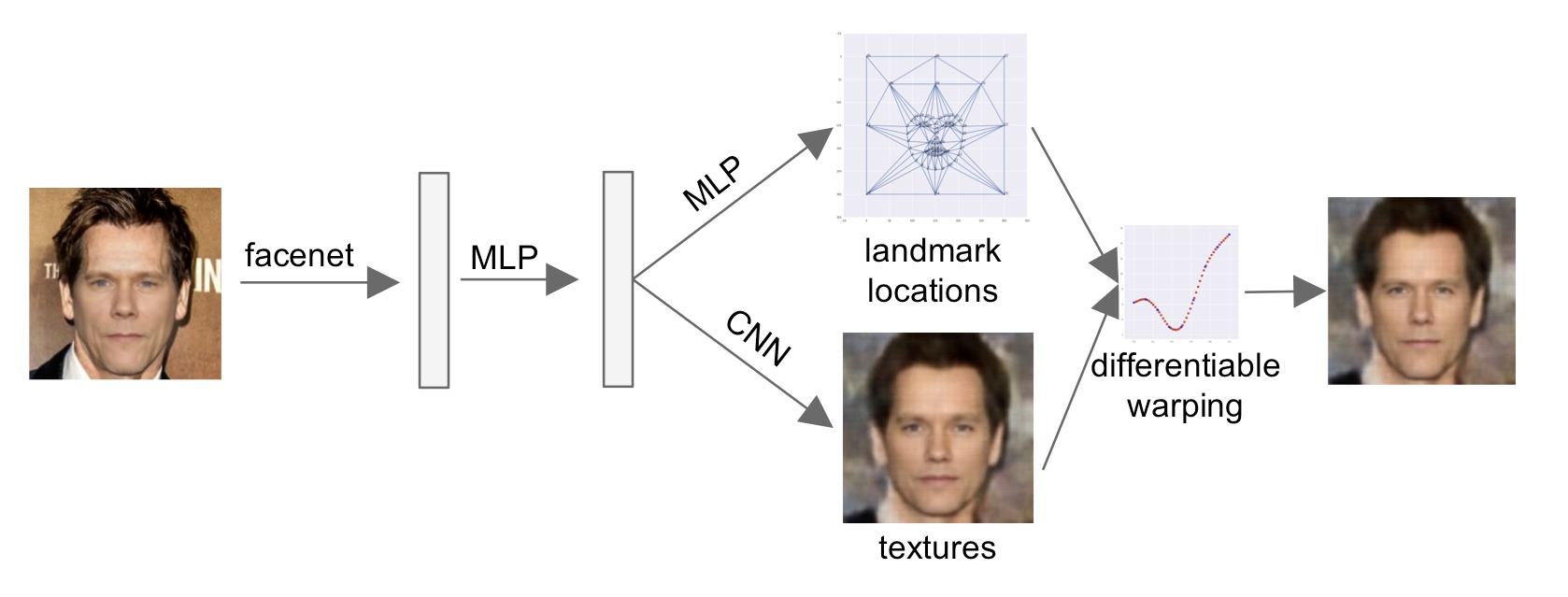}
    \caption{Model Architecture: We first encode an image as a small feature vector using FaceNet~\cite{schroff2015facenet} (with fixed weights) plus an additional multi-layer perceptron (MLP) layer, i.e. a fully connected layer with ReLu non-linearities. Then, we separately generate a texture map, using a deep convolutional network (CNN), and vector of the landmarks' locations, using an MLP. These are combined using differentiable warping to yield the final rendered image.}
    \label{fig:architecture}
\end{figure}

At test time, we consider images taken in the wild, with substantially more variation in lighting, pose, etc. For these, applying our training preprocessing pipeline to obtain $L$ and $T$ is inappropriate. Instead, we use a deep architecture to map directly from the image to estimates of $L$ and $T$. The overall architecture of our network is shown in Fig.~\ref{fig:architecture}.

\subsection{Encoder}
Our encoder takes an input image $I$ and returns an $f$-dimensional feature vector $F$. We need to choose the encoder carefully so that is robust to shifts in the domains of images. In response, we employ a pretrained FaceNet model~\cite{schroff2015facenet} and do not update its parameters. Our assumption is that FaceNet normalizes away variation in face images that is not indicative of the identity of the subject. Therefore, the embeddings of the controlled training images get mapped to the same space as those taken in the wild. This allows us to only train on the controlled images. 

Instead of the final FaceNet output, we use the lowest layer that is not spatially varying: the 1024-D ``avgpool'' layer of the ``NN2'' architecture. We train a fully-connected layer from 1024 to $f$ dimensions on top of this layer. When using VGG-Face features, we use the 4096-D ``fc7'' layer.

\subsection{Decoder}
We could have mapped from $F$ to an output image directly using a deep network. This would need to simultaneously model variation in the geometry and textures of faces. As with~\citet{lanitis1995unified}, we have found it substantially more effective to separately generate landmarks $L$ and textures $T$ and render the final result using warping. 

We generate $L$ using a shallow multi-layer perceptron with ReLU non-linearities applied to $F$. To generate the texture images, we use a deep CNN. We first use a fully-connected layer to map from $F$ to $14 \times 14 \times 256$ localized features. Then, we use a set of stacked transposed convolutions~\cite{dumoulin2016guide}, separated by ReLUs, with a kernel width of $5$ and stride of 2 to upsample to $224 \times 224 \times 32$ localized features. The number of channels after the $i^{th}$ transposed convolution is $\mathtt{max}(256 / 2^i, 32)$. Finally, we apply a $1 \times 1$ convolution to yield $224 \times 224 \times 3$ RGB values. 

Because we are generating registered texture images, it is not unreasonable to use a fully-connected network, rather than a deep CNN. This maps from $F$ to $224\times 224 \times 3$ pixel values directly using a linear transformation. Despite the spatial tiling of the CNN, these models have roughly the same number of parameters. We contrast the outputs of these approaches in Sec.~\ref{sec:design-decisions}. 

The decoder combines the textures and landmarks using the differentiable warping technique described in Sec.~\ref{sec:differentiable warping}. With this, the entire mapping from input image to generated image can be trained end-to-end. 

\subsection{Training Loss}
\label{sec:loss}
\begin{figure}[!ht]
    \includegraphics[width=\columnwidth]{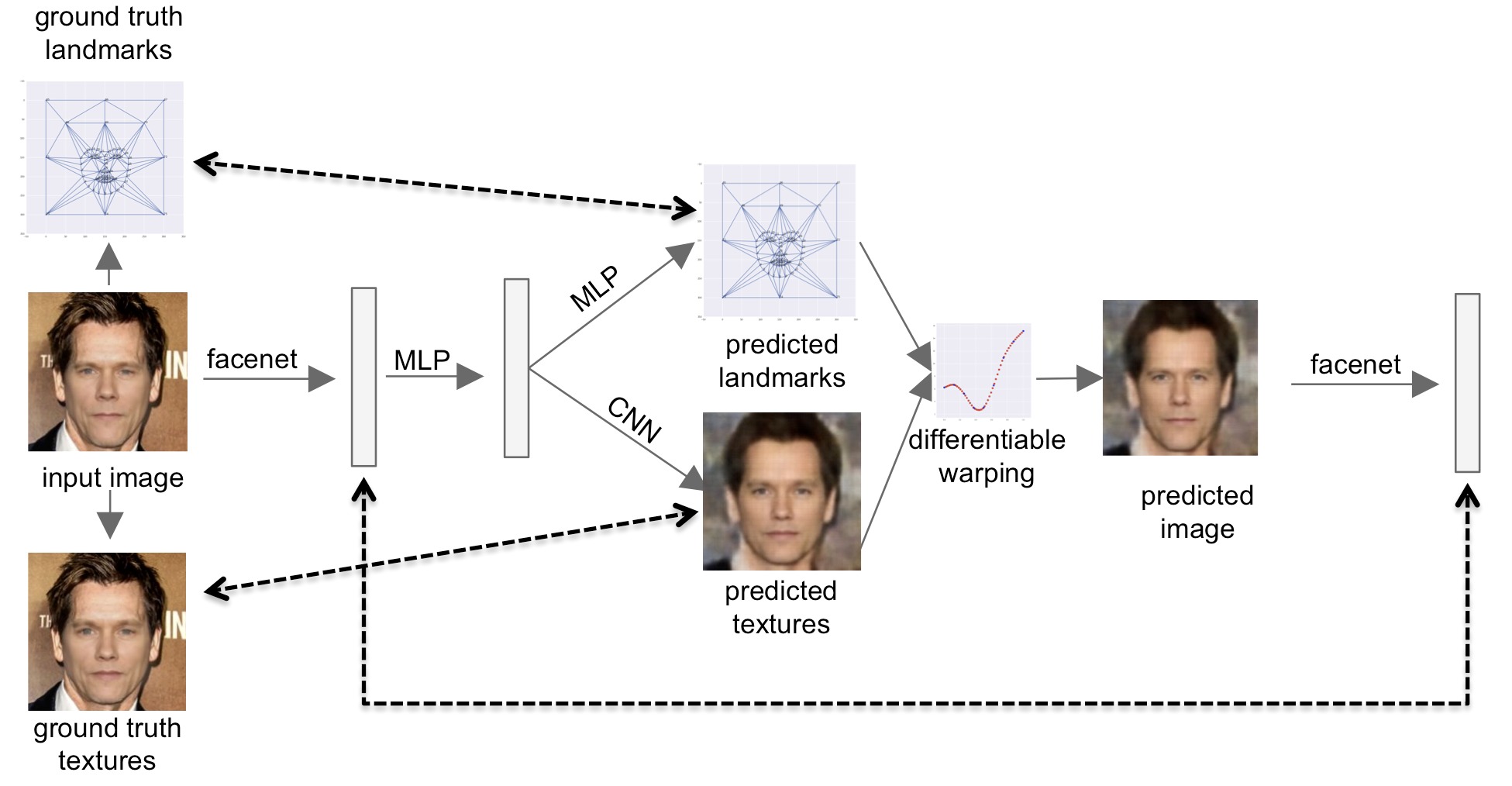}
    \caption{Training Computation Graph: Each dashed line connects two terms that are compared in the loss function. Textures are compared using mean absolute error, landmarks using mean squared error, and FaceNet embedding using negative cosine similarity.}
    \label{fig:training-architecture}
\end{figure}

Our loss function is a sum of the terms depicted in Fig.~\ref{fig:training-architecture}. First, we separately penalize the error of our predicted landmarks and textures, using mean squared error and mean absolute error, respectively. 
This is a more effective loss than penalizing the reconstruction error of the final rendered image. Suppose, for example, that the model predicts the eye color correctly, but the location of the eyes incorrectly. Penalizing reconstruction error of the output image may encourage the eye color to resemble the color of the cheeks. However, by penalizing the landmarks and textures separately, the model will incur no cost for the color prediction, and will only penalize the predicted eye location. 

Next, we reward perceptual similarity between generated images and input images by penalizing the dissimilarity of the FaceNet embeddings of the input and output images. We use a FaceNet network with fixed parameters to compute 128-dimensional embeddings of the two images and penalize their negative cosine similarity. Training with the FaceNet loss adds considerable computational cost: without it, we do not need to perform differentiable warping during training. Furthermore, evaluating FaceNet on the generated image is expensive. 
See Sec.~\ref{sec:design-decisions} for a discussion of the impact of the FaceNet loss on training. 

\section{Differentiable Image Warping}
\label{sec:differentiable warping}
Let $I_0$ be a 2-D image. Let $L = \{(x_1,y_1), \ldots, (x_n, y_n)\}$ be a set of 2-D landmark points and let $D = \{(dx_1,dy_1), \ldots, (dx_n, dy_n)\}$ be a set of displacement vectors for each control point. In the morphable model, $I_0$ is the texture image $T$ and $D = L - \bar{L}$ is the displacement of the landmarks from the mean geometry.  

We seek to warp $I_0$ into a new image $I_1$ such that it satisfies two properties: (a) The landmark points have been shifted by their displacements, i.e. $I_1[x_i,y_i] = I_0[x_i + dx_i, y_i + dy_i]$, and (b) the warping is continuous and resulting flow-field derivatives of any order are controllable. In addition, we require that $I_1$ is a \textit{differentiable} function of $I_0$, $D$, and $L$. We describe our method in terms of 2-D images, but it generalizes naturally to higher dimensions.

\begin{figure}[!ht]
\begin{tabular}{cccc}
  \includegraphics[width=0.2\columnwidth]{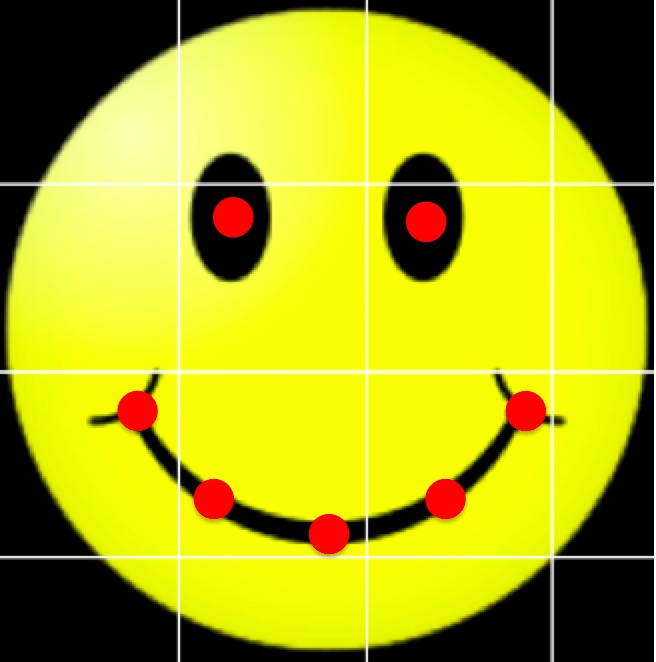} &
  \includegraphics[width=0.2\columnwidth]{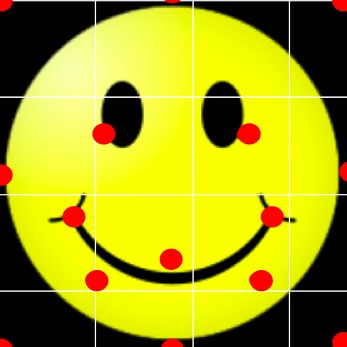} &
   \includegraphics[width=0.2\columnwidth]{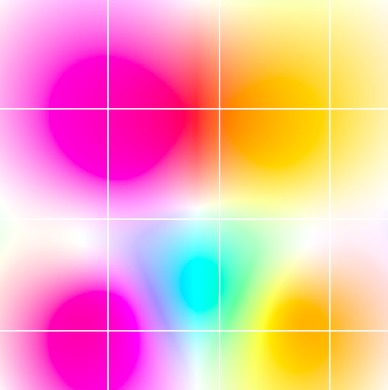} &
  \includegraphics[width=0.2\columnwidth]{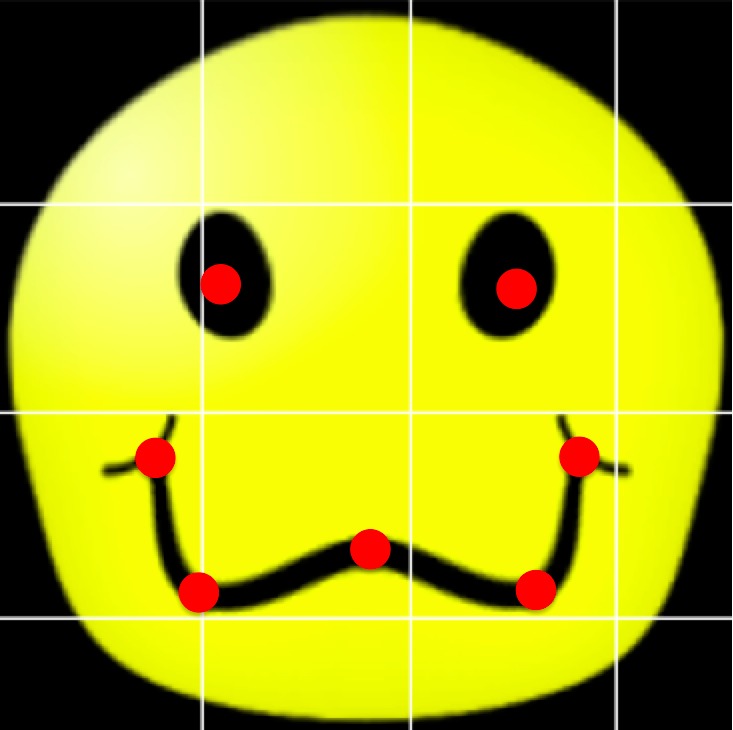} \\
\end{tabular}
\caption{Image warping: Left: starting landmark locations, Middle-left: desired final locations, including zero-displacement boundary conditions, Middle-right: dense flow field obtained by spline interpolation, Right: application of flow to image.}
\label{fig:warp}
\end{figure}

Fig.~\ref{fig:warp} describes our warping. First, we construct a dense flow field from the sparse displacements defined at the control points using spline interpolation. Then, we apply the flow field to $I_0$ in order to obtain $I_1$. The second step uses simple bilinear interpolation, which is differentiable. The next section describes the first step.

\subsection{Differentiable Spline Interpolation}

The interpolation is done independently for horizontal and vertical displacements. For each dimension, we have a scalar $g_p$ defined at each 2-D control point $p$ in $L$ and seek to produce a dense 2-D grid of scalar values. Besides the facial landmark points, we include extra points at the boundary of the image, where we enforce zero displacement. 

We employ polyharmonic interpolation~\cite{iske2012multiresolution}, where the interpolant has the functional form
\begin{equation}
    s(x,y) = \sum_{i = 1}^n w_i \phi_k(\lVert(x,y) - (x_i,y_i) \rVert) + v_1x + v_2y + v_3. \label{eq:spline}
\end{equation}
Here, ${\phi_k}$ are a set of radial basis functions. Common choices are $\phi_1(r) = r$, and $\phi_2(r) = r^2 \log(r)$ (the popular \textit{thin-plate spline}). For our experiments we choose $k=1$, since the linear interpolant is more robust to overshooting than the thin-plate spline, and the linearization artifacts are difficult to detect in the final texture.  

Polyharmonic interpolation chooses the parameters $w_i,a,b,c$ such that $s$ interpolates the signal exactly at the control points, and such that it minimizes a certain definition of curvature~\cite{iske2012multiresolution}. Algorithm~\ref{alg:interp} shows the combined process of estimating the interpolation parameters on training data and evaluating the interpolant at a set of query points. The optimal parameters can be obtained in closed form via operations that are either linear algebra or coordinate-wise non-linearities, all of which are differentiable. Therefore, since~\eqref{eq:spline} is a differentiable function of $x,y$, the entire interpolation process is differentiable.

\RestyleAlgo{boxruled}

\begin{algorithm}[h!]
\small
\centering
   \caption{Differentiable Spline Interpolation}
   \label{alg:mf}
\begin{algorithmic}

   \STATE{\textbf{Inputs:}} points $P = \{(x_1, y_1), \ldots, (x_n, y_n)\}$,\\ function values $G = \{g_1, \ldots, g_n\}$,\\ radial basis function $\phi_k$,\\ query points $Q = \{(x_1, y_1), \ldots, (x_m, y_m)\}$\\
   \STATE{\textbf{Outputs:}} Evaluation of~\eqref{eq:spline} using parameters fit on  $P,F$. 
   \STATE
   \STATE $\text{dists}_{ij} = \lVert P_i - P_j\rVert$ \\
   \STATE $A = \phi_k(\text{dists})$ \\
  \STATE $B = \begin{bmatrix}
    1 & \ldots & 1 \\
    x_1 & \ldots & x_n \\
    y_1 & \ldots & y_n \\
  \end{bmatrix}$ \\
  
 \STATE $\begin{bmatrix}
    w \\
    v  \\
  \end{bmatrix} =  \begin{bmatrix}
    A & B^\top & \\
    B & 0  \\
  \end{bmatrix} \backslash \begin{bmatrix}
    G \\
    0  \\
  \end{bmatrix}$ \% solve linear system \\
  \STATE
  \STATE \textbf{Return} \scalebox{0.95}[1]{$\sum_{i = 1}^n w_i \phi_k(\lVert(x,y) - (x_i,y_i) \rVert)+v_1x+v_2y+ v_3$} 
  \STATE evaluated at each $(x,y)$ point in $Q$.
   
\end{algorithmic}\label{alg:interp}

\end{algorithm}

\section{Data Augmentation using Random Morphs}
\label{sec:data-augmentation}

Training our model requires a large, varied database of evenly-lit, front-facing, neutral-expression photos. Collecting photographs of this type is difficult, and publicly-available databases are too small to train the decoder network (see Fig.~\ref{fig:results-compare}). In response, we construct a small set of high-quality photos and then use a data augmentation approach based on morphing. 

\subsection{Producing random face morphs}


Since the faces are front facing and have similar expressions, we can generate plausible novel faces by morphing. Given a seed face $A$, we first pick a target face by selecting one of the $k = 200$ nearest neighbors of $A$ at random. We measure the distance between faces $A$ and $B$ as:
\begin{equation}
d(A,B) = \lambda \lVert L_A - L_B\rVert + \lVert T_A - T_B\rVert,
\end{equation}
where $L$ are matrices of landmarks and $T$ are texture maps, and $\lambda = 10.0$ in our experiments. Given $A$ and the random neighbor $B$, we linearly interpolate their landmarks and textures independently, where the interpolation weights are drawn uniformly from $[0,1]$. 

\subsection{Gradient-domain Compositing}

\begin{figure}
\begin{tabular}{cccc}
  \includegraphics[width=0.2\columnwidth]{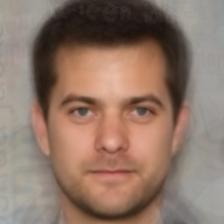} &
  \includegraphics[width=0.2\columnwidth]{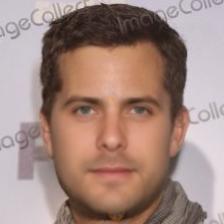} &
  \includegraphics[width=0.2\columnwidth]{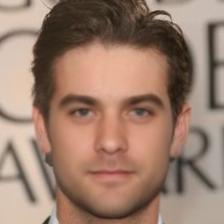} &
  \includegraphics[width=0.2\columnwidth]{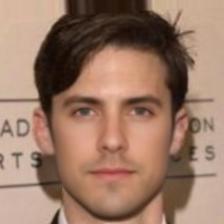} \\
  \includegraphics[width=0.2\columnwidth]{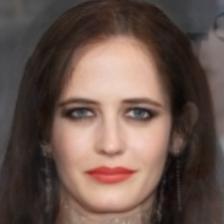} &
  \includegraphics[width=0.2\columnwidth]{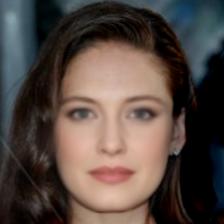} &
  \includegraphics[width=0.2\columnwidth]{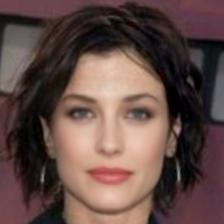} &
  \includegraphics[width=0.2\columnwidth]{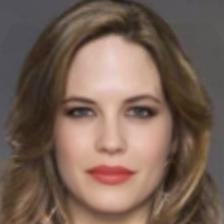} \\
  \includegraphics[width=0.2\columnwidth]{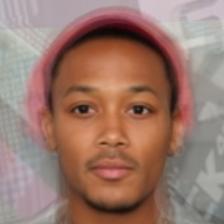} &
  \includegraphics[width=0.2\columnwidth]{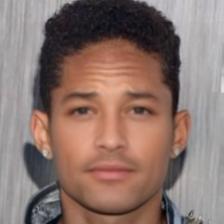} &
  \includegraphics[width=0.2\columnwidth]{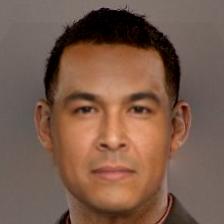} &
  \includegraphics[width=0.2\columnwidth]{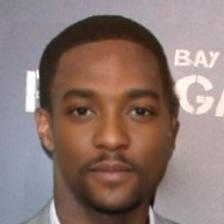} \\
\end{tabular}
\caption{Data augmentation using face morphing and gradient-domain compositing. The left column contains average images of individuals. The remaining columns contain random morphs with  other individuals in the training set.}
\label{fig:training_variations}
\end{figure}

Morphing tends to preserve details inside the face, where the landmarks are accurate, but cannot capture hair and background detail. To make the augmented images more realistic, we paste the morphed face onto an original background using a gradient-domain editing technique \cite{perez2003poisson}. 

Given the texture for a morphed face image $T_f$ and a target background image $T_b$, we construct constraints on the gradient and colors of the output texture $T_o$ as:\begin{equation}
\begin{split}
    \frac{\partial}{\partial x} T_o & = \frac{\partial}{\partial x} T_f \circ M + \frac{\partial}{\partial x} T_b \circ (1 - M) \\
    \frac{\partial}{\partial y} T_o & = \frac{\partial}{\partial y} T_f \circ M + \frac{\partial}{\partial y} T_b \circ (1 - M) \\
    T_o \circ M & = T_f \circ M,
\end{split}
\label{eq:blending}
\end{equation}
where $\circ$ is the element-wise product and the blending mask $M$ is defined by the convex hull of the global average landmarks, softened by a Gaussian blur. Equations~\ref{eq:blending} form an over-constrained linear system that we solve in the least-squares sense. The final result is formed by warping $T_o$ to the morphed landmarks (Fig.~\ref{fig:training_variations}).

\section{Training Data}
\label{sec:training-data}
\subsection{Collecting photographs}

There are a variety of large, publicly-available databases of photographs available online. We choose the dataset used to train the VGG-Face network ~\cite{parkhi2015deep} for its size and its emphasis on facial recognition. It contains 2.6M photographs, but very few of these fit our requirements of front-facing, neutral-pose, and sufficient quality. We use the Google Cloud Vision API~\footnote{cloud.google.com/vision} to remove monochrome and blurry images, faces with high emotion score or eyeglasses, and tilt or pan angles beyond $5^{\circ}$. The remaining images are aligned to undo any roll transformation, scaled to maintain an interocular distance of 55 pixels, and cropped to $224\times224$. After filtering, we have approximately 12K images ($<0.5\%$ of the original set). 

\subsection{Averaging to reduce lighting variation}

\begin{figure}
\begin{tabular}{ccc|c}
  & Inputs & & Averaged \\
  \includegraphics[width=0.2\columnwidth]{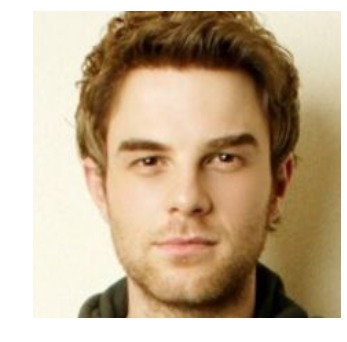} &
  \includegraphics[width=0.2\columnwidth]{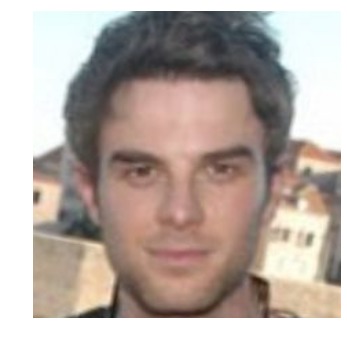} &
  \includegraphics[width=0.2\columnwidth]{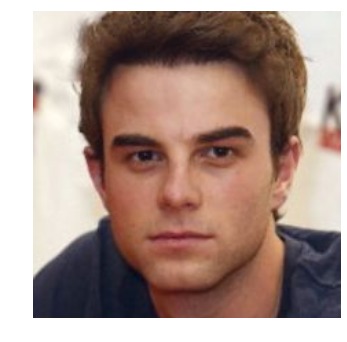} &
  \includegraphics[width=0.2\columnwidth]{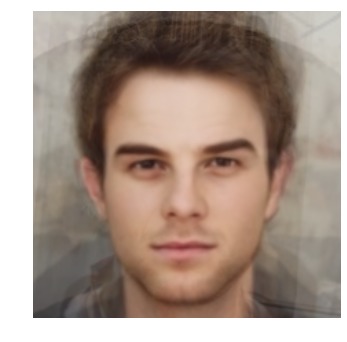} \\
\end{tabular}
\caption{Averaging images of the same individual to produce consistent lighting. Example input photographs (left three columns) have large variation in lighting and color. Averaging tends to produce an evenly lit, but still detailed, result (right column). }
\label{fig:average_morph}
\end{figure}

To further remove variation in lighting, we average all images for each individual by morphing. After filtering for quality, we have $\approx$1K unique identities that have 3 or more images per identity. Given the set of images of an individual $\mathcal{I}_j$, we extract facial landmarks  $L_j$ for each image using the method of Kazemi and Sullivan \cite{kazemi2014onemilli} and then average the landmarks to form $L_{\mu}$. Each image $\mathcal{I}_j$ is warped to the average landmarks $L_{\mu}$, then the pixel values are averaged to form an average image of the individual $\mathcal{I}_{\mu}$. As shown in Fig.~\ref{fig:average_morph}, this operation tends to produce images that resemble photographs with soft, even lighting. These 1K images form the base training set.

The backgrounds in the training images are widely variable, leading to noisy backgrounds in our results. Cleaner results could probably be obtained by manual removal of the backgrounds.

\section{Experiments}
\label{sec:experiments}
For our experiments we mainly focus on the Labeled Faces in the Wild~\cite{huang2007lfw} dataset, since its identities are mutually exclusive with the VGG face dataset. We include a few example from other sources, such as a painting, to show the range of the method.

Except where otherwise noted, the results were produced with the architecture of Section~\ref{sec:model}, with weights on the landmark loss $=1$, the FaceNet loss $=10$, and texture loss $=100$. Our data augmentation produces 1M images. The model was implemented in TensorFlow~\cite{tensorflow2015-whitepaper} and trained using the Adam optimizer~\cite{kingma2014adam}. 

\subsection{Model Robustness}

\begin{figure*}
\begin{tabular}{ccccc|cccc}

  \hspace{-1.0ex}Input &
 \includegraphics[align=c,width=0.21\columnwidth]{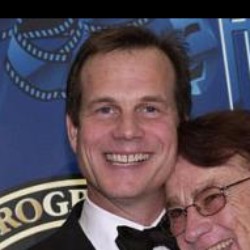} &
  \hspace{-1.0ex}\includegraphics[align=c,width=0.21\columnwidth]{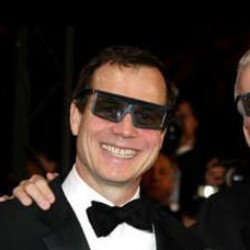} &
  \hspace{-1.0ex}\includegraphics[align=c,width=0.21\columnwidth]{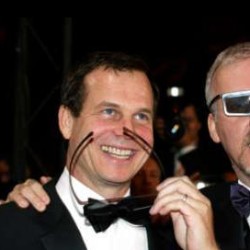} &
  \hspace{-1.0ex}\includegraphics[align=c,width=0.21\columnwidth]{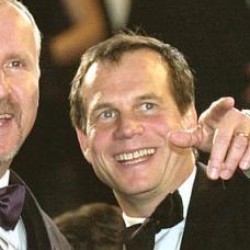} &
  \includegraphics[align=c,width=0.21\columnwidth]{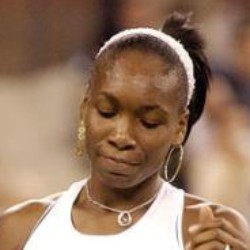} & 
  \hspace{-1.0ex}\includegraphics[align=c,width=0.21\columnwidth]{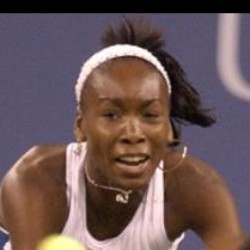} &
  \hspace{-1.0ex}\includegraphics[align=c,width=0.21\columnwidth]{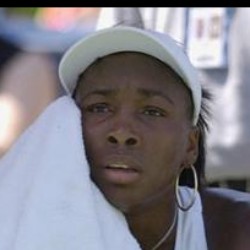} &
  \hspace{-1.0ex}\includegraphics[align=c,width=0.21\columnwidth]{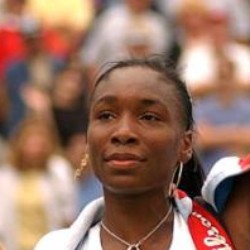} \vspace{3.0pt} \\

  \hspace{-1.0ex}FN &
  \includegraphics[align=c,width=0.21\columnwidth]{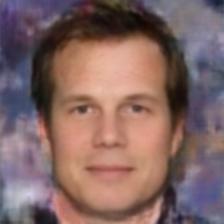} & 
  \hspace{-1.0ex}\includegraphics[align=c,width=0.21\columnwidth]{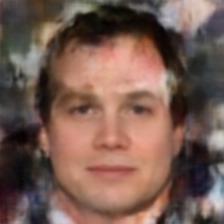} &
  \hspace{-1.0ex}\includegraphics[align=c,width=0.21\columnwidth]{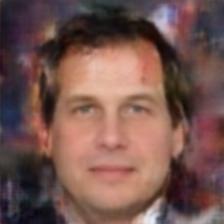} &
  \hspace{-1.0ex}\includegraphics[align=c,width=0.21\columnwidth]{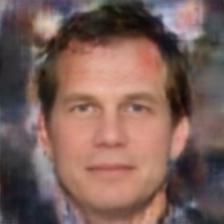} &
  \includegraphics[align=c,width=0.21\columnwidth]{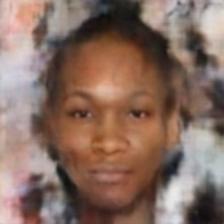} & 
  \hspace{-1.0ex}\includegraphics[align=c,width=0.21\columnwidth]{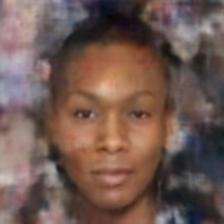} &
  \hspace{-1.0ex}\includegraphics[align=c,width=0.21\columnwidth]{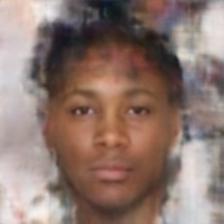} &
  \hspace{-1.0ex}\includegraphics[align=c,width=0.21\columnwidth]{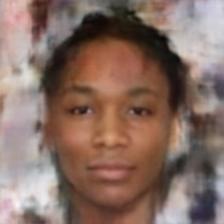} \vspace{3.0pt} \\

  \hspace{-1.0ex}VGG &
  \includegraphics[align=c,width=0.21\columnwidth]{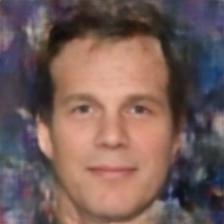} & 
  \hspace{-1.0ex}\includegraphics[align=c,width=0.21\columnwidth]{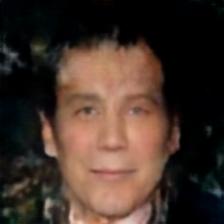} &
  \hspace{-1.0ex}\includegraphics[align=c,width=0.21\columnwidth]{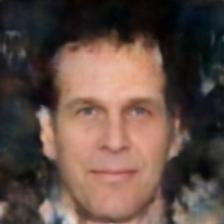} &
  \hspace{-1.0ex}\includegraphics[align=c,width=0.21\columnwidth]{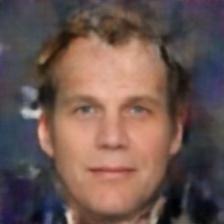} &
  \includegraphics[align=c,width=0.21\columnwidth]{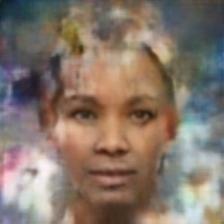} & 
  \hspace{-1.0ex}\includegraphics[align=c,width=0.21\columnwidth]{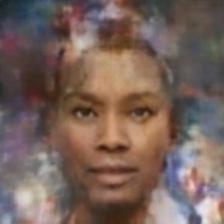} &
  \hspace{-1.0ex}\includegraphics[align=c,width=0.21\columnwidth]{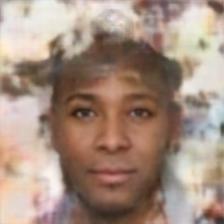} &
  \hspace{-1.0ex}\includegraphics[align=c,width=0.21\columnwidth]{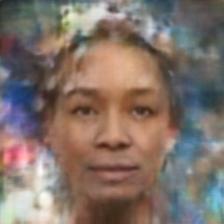} \vspace{3.0pt} \\
  
  \hspace{-1.0ex}\cite{hassner2015effective} &
  \includegraphics[align=c,width=0.11\columnwidth]{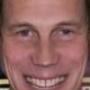} & 
  \hspace{-1.0ex}\includegraphics[align=c,width=0.11\columnwidth]{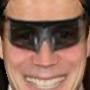} &
  \hspace{-1.0ex}\includegraphics[align=c,width=0.11\columnwidth]{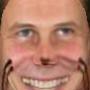} &
  \hspace{-1.0ex}\includegraphics[align=c,width=0.11\columnwidth]{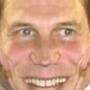} &
  \includegraphics[align=c,width=0.11\columnwidth]{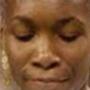} & 
  \hspace{-1.0ex}\includegraphics[align=c,width=0.11\columnwidth]{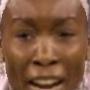} &
  \hspace{-1.0ex}\includegraphics[align=c,width=0.11\columnwidth]{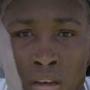} &
  \hspace{-1.0ex}\includegraphics[align=c,width=0.11\columnwidth]{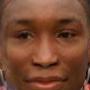} \vspace{3.0pt} \\
  
\end{tabular}
\vspace{-0.5ex}
\caption{Face normalization for people in the LFW dataset \cite{huang2007lfw}. Top to bottom: input photographs, result of our method using FaceNet features, result of our method using VGG-Face features,  result of Hassner, et al. \cite{hassner2015effective}. 
Additional results in supplementary material.} \label{fig:frontalization}
\end{figure*}

Fig.~\ref{fig:frontalization} shows the robustness of our model to nuisance factors such as occlusion, pose and illumination. We use two identities from the LFW dataset~\citep{huang2007lfw}, and four images for each identity (top row). Our model's results when trained on FaceNet ``avgpool-0'' and VGG-Face ``fc7'' features are shown in middle rows. The results from the FaceNet features are especially stable across different poses and illumination, but the VGG-Face features are comparable. Severe occlusions such as sunglasses and headwear do not significantly impact the output quality. The model even works on paintings, such as Fig.~\ref{fig:intro-examples} (right) and Fig.~\ref{fig:3dresults} (top right).

For comparison, we include a state-of-the-art frontalization method based on image warping (\citet{hassner2015effective}). In contrast to our method, image warping does not remove occlusions, handle extreme poses,  neutralize some expressions, or correct for variability in illumination. 

\subsection{Impact of Design Decisions}
\label{sec:design-decisions}

\addtolength{\tabcolsep}{-4pt}    
\begin{figure}
\begin{tabular}{cccc}
 & {\smaller CNN w/o Data Aug.}  & {\smaller FC w/ Data Aug.}  & {\smaller CNN w/ Data Aug.} \\ 
  
 \includegraphics[width=0.18\columnwidth]{images/frontalization/original/Laura_Bush_0001} &
 \includegraphics[width=0.18\columnwidth]{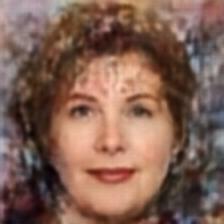} &
 \includegraphics[width=0.18\columnwidth]{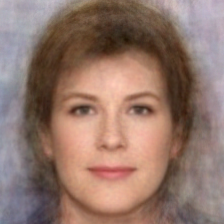} &
 \includegraphics[width=0.18\columnwidth]{images/frontalization/ours/Laura_Bush_0001} \\ 
  \includegraphics[width=0.18\columnwidth]{images/frontalization/original/Vince_Carter_0001} &
 \includegraphics[width=0.18\columnwidth]{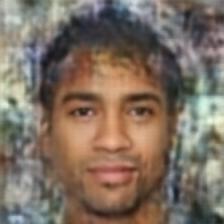} &
 \includegraphics[width=0.18\columnwidth]{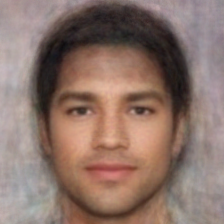} &
 \includegraphics[width=0.18\columnwidth]{images/frontalization/ours/Vince_Carter_0001}
\end{tabular}
\caption{Output from various configurations of our system: CNN texture decoder trained with only 1K raw images, fully-connected decoder and CNN trained on 1M images using the data augmentation technique of Sec.~\ref{sec:data-augmentation}. } 
\label{fig:results-compare}
\end{figure}
\addtolength{\tabcolsep}{4pt}

In Fig.~\ref{fig:results-compare} we contrast the output of our system with two variations: a model trained without data augmentation and a model that uses data augmentation, but employs a fully-connected network for predicting textures. Training without data augmentation yields more artifacts due to overfitting. The fully-connected decoder generates images that are very generic, since though it has separate parameters for every pixel, its capacity is limited because there is no mechanism for coordinating outputs at multiple scales. 

\addtolength{\tabcolsep}{-2pt}    
\begin{figure}[t]
\vspace{-1em}
\begin{center}
\begin{tabular}{ccc}
   Input & Plain CNN & Our method \\
   \includegraphics[width=0.3\linewidth]{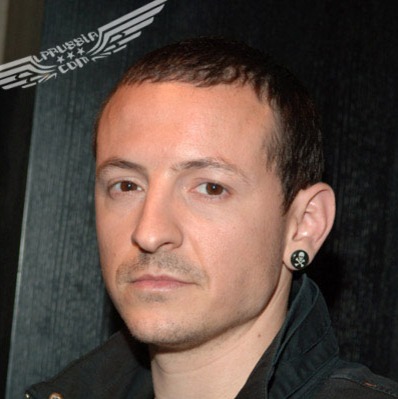} &
   \includegraphics[width=0.3\linewidth]{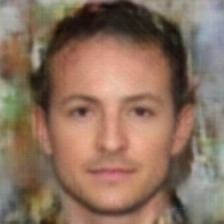} &
   \includegraphics[width=0.3\linewidth]{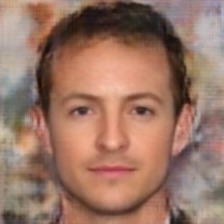} \\
   &
   \includegraphics[width=0.3\linewidth]{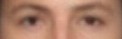} &
   \includegraphics[width=0.3\linewidth]{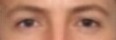} \\

\end{tabular}
\end{center}
\vspace{-1em}
   \caption{Decoder architecture comparison on test data. ``Plain CNN'' does not decouple texture and landmarks, while our method does. Decoder capacities and training regime are identical.}
\label{input_figure_compare}
\vspace{-1.5em}
\end{figure}
\addtolength{\tabcolsep}{2pt}

Fig.~\ref{input_figure_compare} shows the benefit of decoupling texture and landmark prediction. Compared to a regular CNN with the same decoder capacity, our method reproduces finer details. The increased performance results from the main observation of \citet{lanitis1995unified}: warping the input images to the global mean landmarks (Fig. 2) aligns features such as eyes and lips across the training set, allowing the decoder to fit the face images with higher fidelity. 

\begin{figure}
\begin{tabular}{ccc}
 & {\smaller w/ FaceNet loss} & {\smaller w/o FaceNet loss} \\
 {\smaller Input } & {\smaller FN $L_2$ error : 0.42} & {\smaller FN $L_2$ error: 0.8} \\
  \includegraphics[width=0.28\columnwidth]{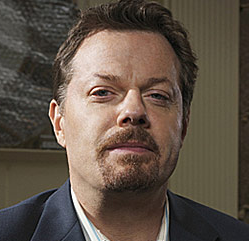} &
  \includegraphics[width=0.28\columnwidth]{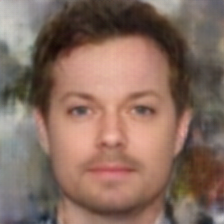} &
  \includegraphics[width=0.28\columnwidth]{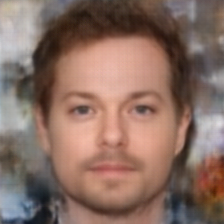} 
\end{tabular}
\caption{Results with and without loss term penalizing difference in the FaceNet embedding. The FaceNet loss encourages subtle but important improvements in fidelity, especially around the eyes and eyebrows. The result is a lower error between the embeddings of the input and synthesized images.}
\label{fig:facenet_with_without}
\end{figure}

Fig.~\ref{fig:facenet_with_without} compares outputs of models trained with and without the FaceNet loss. The difference is subtle but visible, and has a perceptual effect of improving the likeness of the recovered image.

\begin{figure}
\begin{center}
\includegraphics[width=0.85\columnwidth]{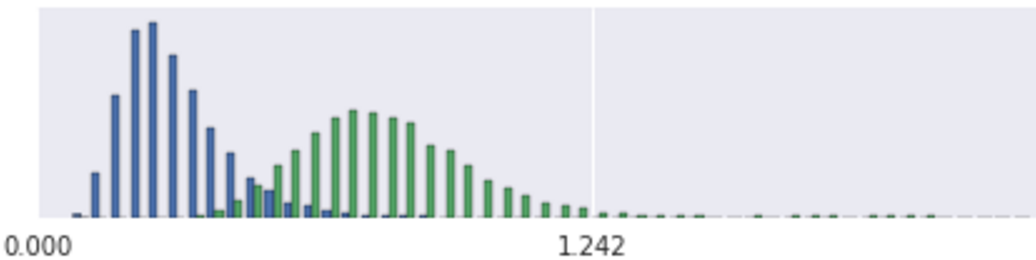}
\end{center}
  
\caption{Histograms of FaceNet $L_2$ error between input and synthesized images on LFW. Blue: with FaceNet loss (Sec.~\ref{sec:loss}). Green: without FaceNet loss. The $1.242$ threshold was used by \citet{schroff2015facenet} to cluster identities. Without the FaceNet loss, about 2\% of the synthesized images would not be considered the same identity as the input image.}
\label{fig:facenet_histogram}
\end{figure}

The improvement from training with the FaceNet loss can also be measured by evaluating FaceNet on the test outputs. Fig.~\ref{fig:facenet_histogram} shows the distributions of $L_2$ distances between the embeddings of the LFW images and their corresponding synthesized results, for models trained with and without the FaceNet loss.
\citet{schroff2015facenet} consider two FaceNet embeddings to encode the same person if their $L_2$ distance is less than $1.242$. All of the synthesized images pass this test using FaceNet loss, but without, about 2\% of the images would be mid-identified by FaceNet as a different person.



\subsection{3-D Model Fitting}
\label{sec:3d_model_fitting}

\begin{figure}
\includegraphics[width=\columnwidth]{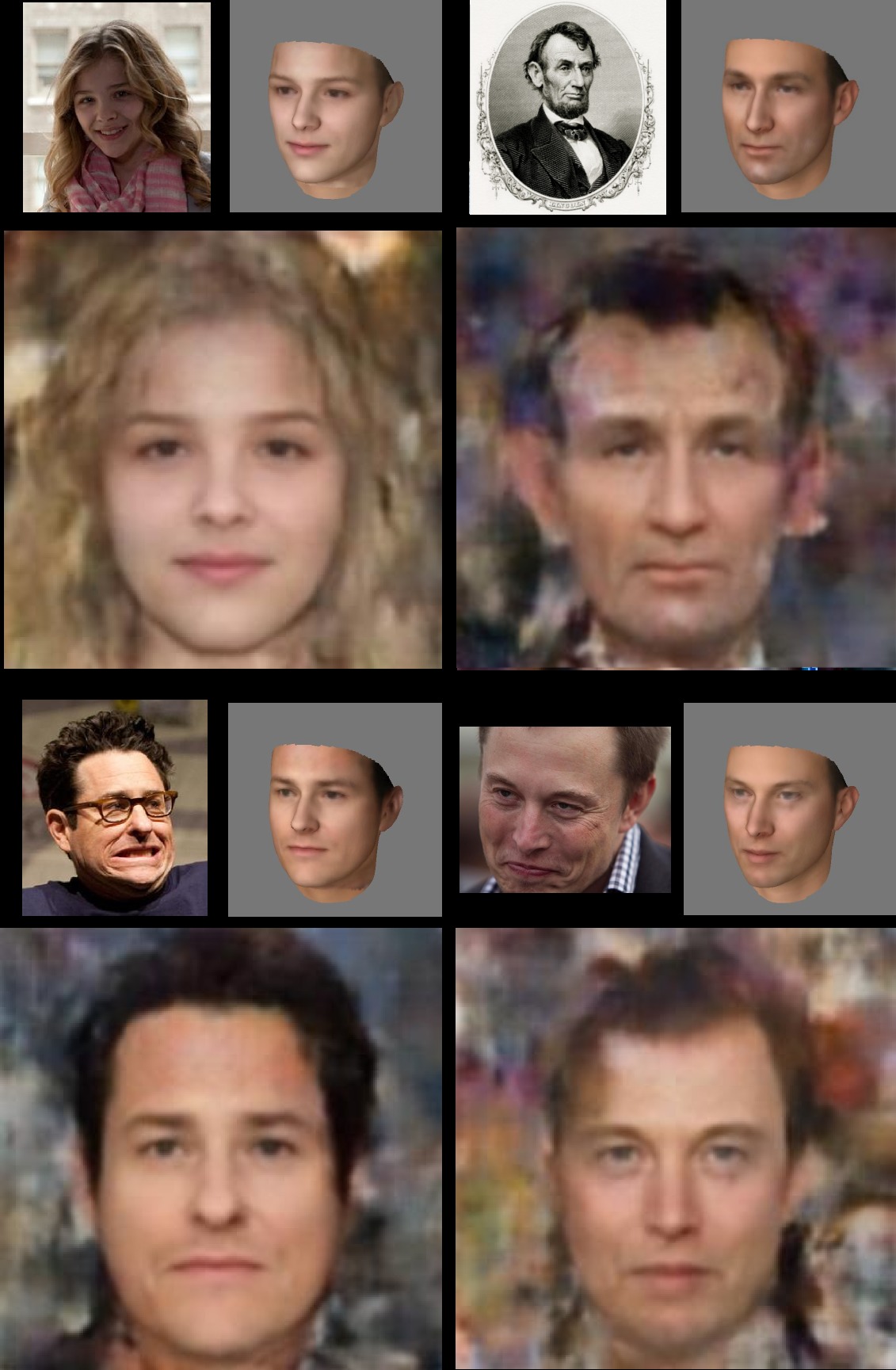}
\caption{Mapping of our model's output onto a 3-D face. Small: input and fit 3-D model. Large: synthesized 2-D image. Photos by Wired.com, CC BY-NC 2.0 (images were cropped).}
\label{fig:3dresults}
\end{figure}

The landmarks and texture of the normalized face can be used to fit a 3D morphable model (Fig.~\ref{fig:3dresults}). Fitting a morphable model to an unconstrained image of a face requires solving a difficult inverse rendering problem~\cite{blanz1999morphable}, but fitting to a normalized face image is much more straightforward. See Sec. 2 of the supplementary material for details. 

The process produces a well-aligned, 3D face mesh that could be directly used as a VR avatar, or could serve as an initialization for further processing, for example in methods to track facial geometry in video~\cite{suwajanakorn2014total,suwajanakorn2015}. 
The fidelity of the reconstructed shape is limited by the range of the morphable model, and could likely be improved with a more diverse model such as the recent LSFM~\cite{booth2016lsfm}.



\subsection{Automatic Photo Adjustment}
\label{sec:whitebalance}

\begin{figure}
\vspace{-1ex}
\begin{tabular}{cccc}
  & \multicolumn{2}{c}{Input Images} &\\
  \includegraphics[width=0.2\columnwidth]{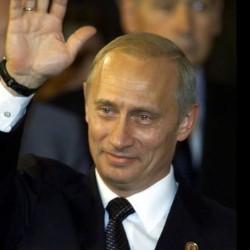} &
  \includegraphics[width=0.2\columnwidth]{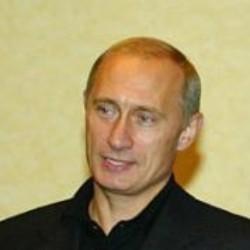} &
  \includegraphics[width=0.2\columnwidth]{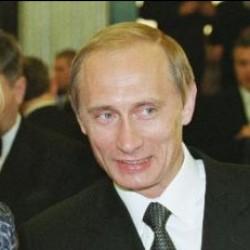} &
  \includegraphics[width=0.2\columnwidth]{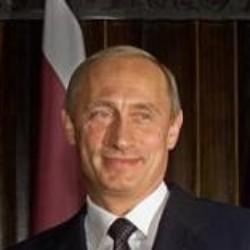} \\
  & \multicolumn{2}{c}{Our Method} &\\
  \includegraphics[width=0.2\columnwidth]{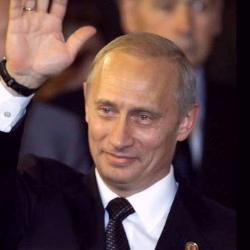} &
  \includegraphics[width=0.2\columnwidth]{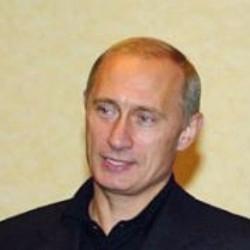} &
  \includegraphics[width=0.2\columnwidth]{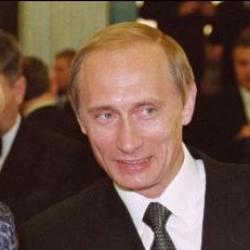} &
  \includegraphics[width=0.2\columnwidth]{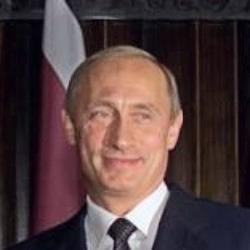} \\
  & \multicolumn{2}{c}{Barron \cite{barron2015color}} &\\
  \includegraphics[width=0.2\columnwidth]{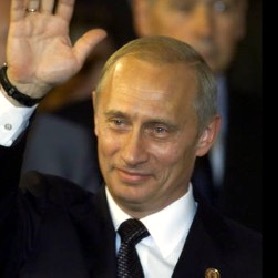} &
  \includegraphics[width=0.2\columnwidth]{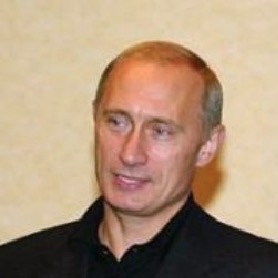} &
  \includegraphics[width=0.2\columnwidth]{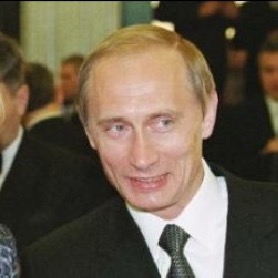} &
  \includegraphics[width=0.2\columnwidth]{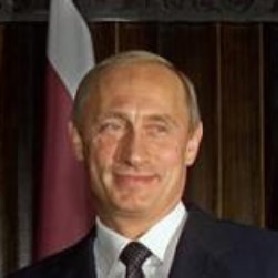} \\
  & \multicolumn{2}{c}{Input Images} &\\
  \includegraphics[width=0.2\columnwidth]{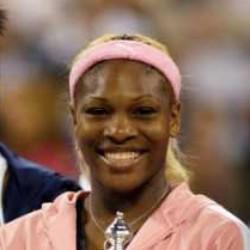} & 
  \includegraphics[width=0.2\columnwidth]{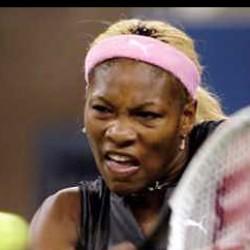} & 
  \includegraphics[width=0.2\columnwidth]{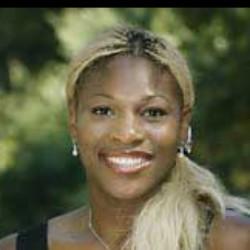} &
  \includegraphics[width=0.2\columnwidth]{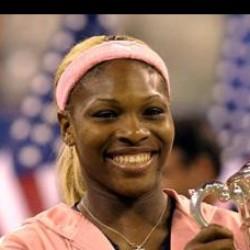} \\
  & \multicolumn{2}{c}{Our Method} &\\
  \includegraphics[width=0.2\columnwidth]{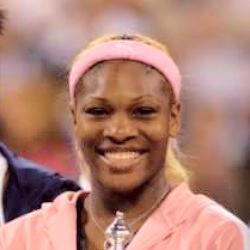} & 
  \includegraphics[width=0.2\columnwidth]{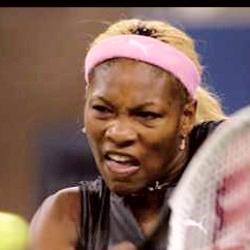} & 
  \includegraphics[width=0.2\columnwidth]{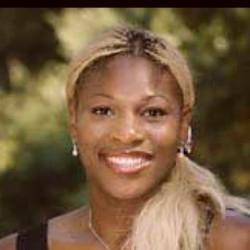} &
  \includegraphics[width=0.2\columnwidth]{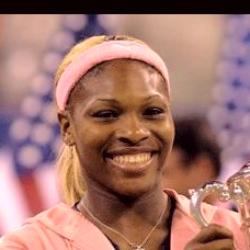} \\
  & \multicolumn{2}{c}{Barron \cite{barron2015color}} &\\
  \includegraphics[width=0.2\columnwidth]{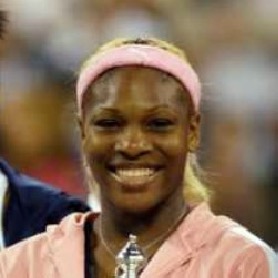} & 
  \includegraphics[width=0.2\columnwidth]{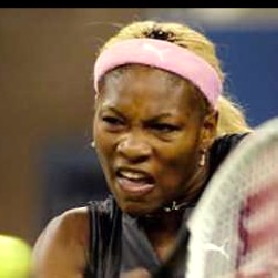} & 
  \includegraphics[width=0.2\columnwidth]{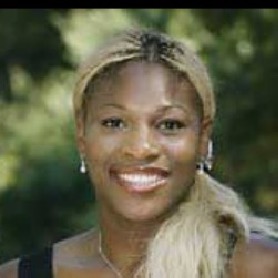} &
  \includegraphics[width=0.2\columnwidth]{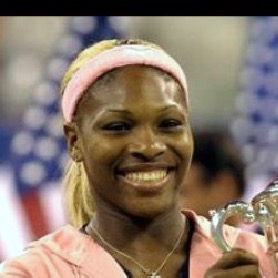} \\
\end{tabular}
\caption{Automatic adjustment of exposure and white balance using the color of the normalized face for some images from the LFW dataset. In each set of images (2 sets of 3 rows), the first row are the input images; the second row the outputs from out method and the third row the outputs of Barron~\cite{barron2015color}, a state-of-the-art white balancing method. The implicit encoding of skin tone in our model is crucial to the exposure and white balance recovery.}
\label{fig:whitebalance}
\end{figure}

Since the normalized face image provides a ``ground truth'' image of the face, it can be easily applied to automatically adjust the exposure and white balance of a photograph (Fig.~\ref{fig:whitebalance}). We apply the following simple algorithm: given an aligned input photograph $P$ and the corresponding normalized face image $N$, extract a box from the center of $P$ and $N$ (in our experiments, the central $100\times100$ pixels out of $224\times224$) and average the cropped regions to form mean face colors $\mathbf{m}_P$ and $\mathbf{m}_N$. The adjusted image is computed using a per-channel, piecewise-linear color shift function. See Sec. 3 of the supplementary material for details. 

For comparison, we apply the general white balancing algorithm of~\citet{barron2015color}. This approach does not focus on the face, and is limited in the adjustment it makes, whereas our algorithm balances the face regardless of the effect on the other regions of the image, producing more consistent results across different photos of the same person.




\section{Conclusion and Future Work}
\label{sec:conclusion}
We have introduced a neural network that maps from images of faces taken in the wild to front-facing neutral-expression images that  capture the likeness of the individual. The network is robust to variation in the inputs, such as lighting, pose, and expression, that cause problems for prior face frontalization methods. The method provides a variety of down-stream opportunities, including  automatically white-balancing images and creating custom 3-D avatars. 

Spline interpolation has been used extensively in computer graphics, but we are unaware of work where interpolation has been used as a differentiable module inside a network. We encourage further application of the technique. 

We hope to improve our images' quality. 
Noise artifacts likely result from overfitting to the images' backgrounds and blurriness likely results from using a pixel-level squared error. Ideally, we would use a broad selection of training images and avoid pixel-level losses entirely, by combining the FaceNet loss of Sec.~\ref{sec:loss} with an adversarial loss~\citep{goodfellow2014generative}.
{\small
\bibliographystyle{IEEEtranN}
\bibliography{sources}
}

\appendix

\section{Additional Results}

Figures~\ref{fig:frontalization_supplemental} and ~\ref{fig:frontalization2} contain additional results of face normalization on LFW and comparison to \citet{hassner2015effective}. 

Figure~\ref{fig:3dresults_supplemental} show results from degraded photographs and illustrations, which push the method outside of its training domain but still produce credible results.

\section{3-D Model Fitting}
\label{sec:3d_model_fitting_supplemental}

To fit the shape of the face, we first manually establish a correspondence between the 65 predicted landmarks $\mathbf{l}_i$ and the best matching 65 vertices $\mathbf{v}_i$ of the 3-D mesh used to train the model of~\citet{blanz1999morphable}. This correspondence is based on the semantics of the landmarks and does not change for different faces. We then optimize for the shape parameters that best match $\mathbf{v}_i$ to $\mathbf{l}_i$ using gradient descent. The landmarks provide $65\times2=130$ constraints for the $199$ parameters of the morphable model, so the optimization is additionally regularized towards the average face. 

Once the face mesh is aligned with the predicted landmarks, we project the synthesized image onto the mesh as vertex colors. The projection works well for areas that are close to front-facing, but is noisy and imprecise at grazing angles. To clean the result, we project the colors further onto the model's texture basis to produce clean, but less accurate vertex colors. We then produce a final vertex color by blending the synthesized image color and the texture basis color based on the foreshortening angle. 

\subsection{Corresponding Landmarks and Vertices}

As a pre-processing step, we determine which 65 vertices of the shape model's mesh best match the 65 landmark positions. Since the topology of the mesh doesn't change as the shape changes, the correspondence between landmark indices and vertex indices is fixed. 

\begin{figure}[h]
    \includegraphics[width=0.5\columnwidth]{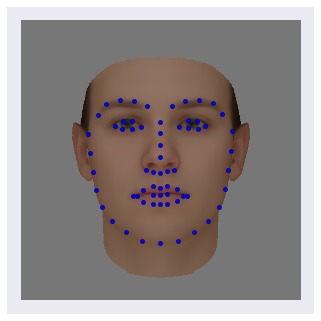}
    \caption{Landmarks extracted from the mean face of the Blanz and Vetter model.}
    \label{fig:mean-face-landmarks}
\end{figure}

\begin{figure*}
\vspace{-2em}
\begin{tabular}{ccccc|cccc}

 \hspace{-1.0ex}Input &
 \includegraphics[align=c,width=0.21\columnwidth]{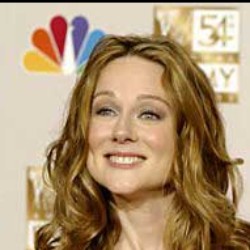} &
  \hspace{-1.0ex}\includegraphics[align=c,width=0.21\columnwidth]{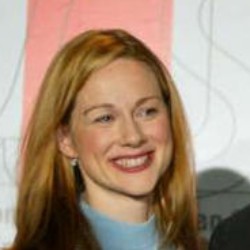} &
  \hspace{-1.0ex}\includegraphics[align=c,width=0.21\columnwidth]{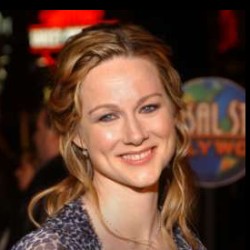} &
  \hspace{-1.0ex}\includegraphics[align=c,width=0.21\columnwidth]{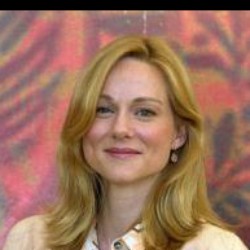} &
  \includegraphics[align=c,width=0.21\columnwidth]{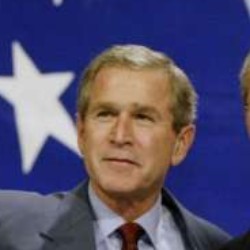} & 
  \hspace{-1.0ex}\includegraphics[align=c,width=0.21\columnwidth]{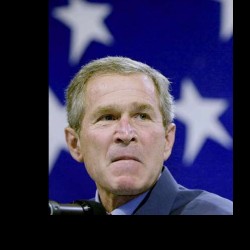} &
  \hspace{-1.0ex}\includegraphics[align=c,width=0.21\columnwidth]{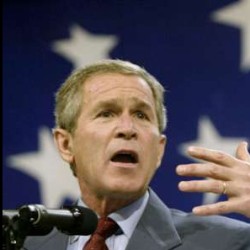} &
  \hspace{-1.0ex}\includegraphics[align=c,width=0.21\columnwidth]{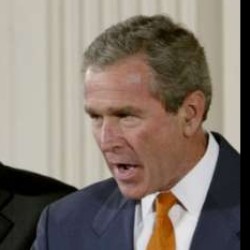} \vspace{3.0pt} \\

  \hspace{-1.0ex}FN &
  \includegraphics[align=c,width=0.21\columnwidth]{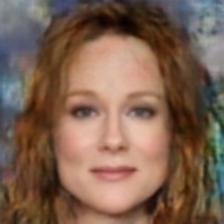} & 
  \hspace{-1.0ex}\includegraphics[align=c,width=0.21\columnwidth]{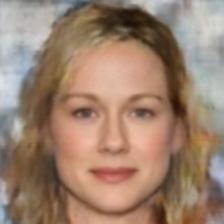} &
  \hspace{-1.0ex}\includegraphics[align=c,width=0.21\columnwidth]{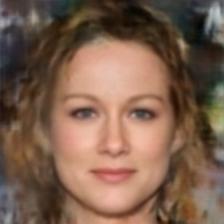} &
  \hspace{-1.0ex}\includegraphics[align=c,width=0.21\columnwidth]{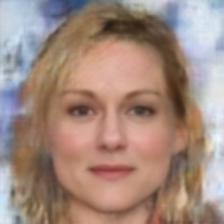} &
  \includegraphics[align=c,width=0.21\columnwidth]{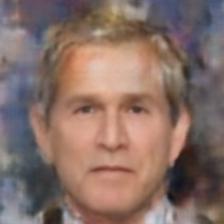} & 
  \hspace{-1.0ex}\includegraphics[align=c,width=0.21\columnwidth]{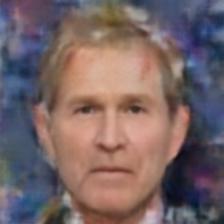} &
  \hspace{-1.0ex}\includegraphics[align=c,width=0.21\columnwidth]{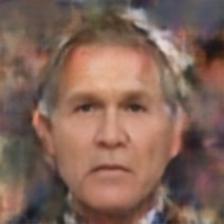} &
  \hspace{-1.0ex}\includegraphics[align=c,width=0.21\columnwidth]{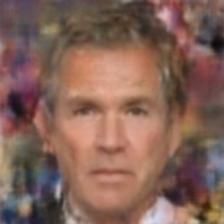} \vspace{3.0pt} \\

  \hspace{-1.0ex}VGG &
  \includegraphics[align=c,width=0.21\columnwidth]{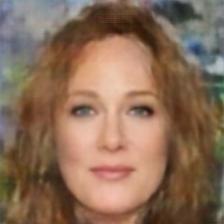} & 
  \hspace{-1.0ex}\includegraphics[align=c,width=0.21\columnwidth]{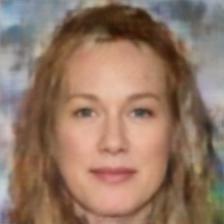} &
  \hspace{-1.0ex}\includegraphics[align=c,width=0.21\columnwidth]{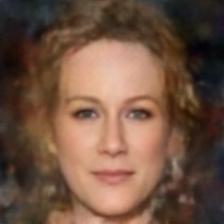} &
  \hspace{-1.0ex}\includegraphics[align=c,width=0.21\columnwidth]{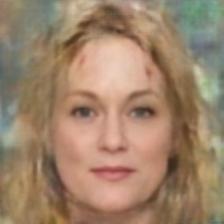} &
  \includegraphics[align=c,width=0.21\columnwidth]{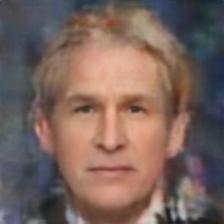} & 
  \hspace{-1.0ex}\includegraphics[align=c,width=0.21\columnwidth]{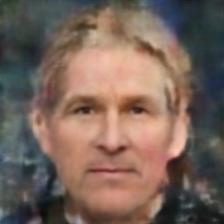} &
  \hspace{-1.0ex}\includegraphics[align=c,width=0.21\columnwidth]{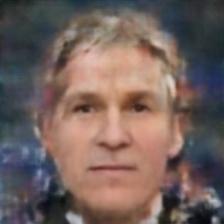} &
  \hspace{-1.0ex}\includegraphics[align=c,width=0.21\columnwidth]{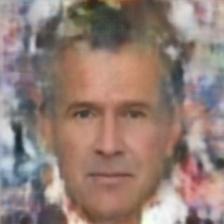} \vspace{3.0pt} \\
  
  \hspace{-1.0ex}\cite{hassner2015effective} &
  \includegraphics[align=c,width=0.11\columnwidth]{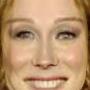} & 
  \hspace{-1.0ex}\includegraphics[align=c,width=0.11\columnwidth]{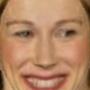} &
  \hspace{-1.0ex}\includegraphics[align=c,width=0.11\columnwidth]{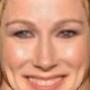} &
  \hspace{-1.0ex}\includegraphics[align=c,width=0.11\columnwidth]{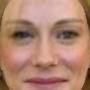} &
  \includegraphics[align=c,width=0.11\columnwidth]{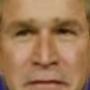} & 
  \hspace{-1.0ex}\includegraphics[align=c,width=0.11\columnwidth]{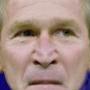} &
  \hspace{-1.0ex}\includegraphics[align=c,width=0.11\columnwidth]{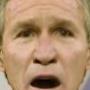} &
  \hspace{-1.0ex}\includegraphics[align=c,width=0.11\columnwidth]{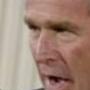} \vspace{3.0pt} \\
  
  \hline

   \hspace{-1.0ex}Input &
 \includegraphics[align=c,width=0.21\columnwidth]{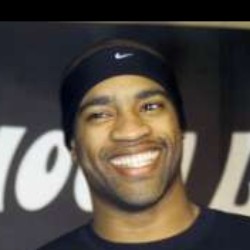} &
  \hspace{-1.0ex}\includegraphics[align=c,width=0.21\columnwidth]{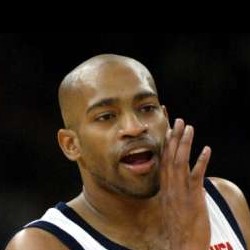} &
  \hspace{-1.0ex}\includegraphics[align=c,width=0.21\columnwidth]{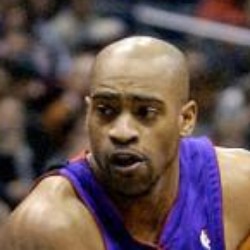} &
  \hspace{-1.0ex}\includegraphics[align=c,width=0.21\columnwidth]{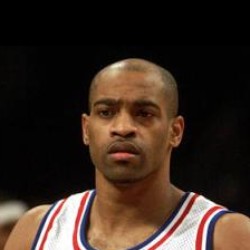} &
  \includegraphics[align=c,width=0.21\columnwidth]{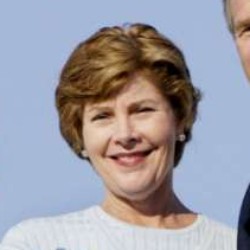} & 
  \hspace{-1.0ex}\includegraphics[align=c,width=0.21\columnwidth]{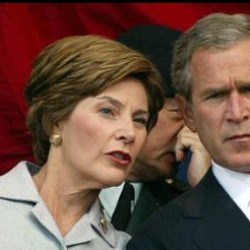} &
  \hspace{-1.0ex}\includegraphics[align=c,width=0.21\columnwidth]{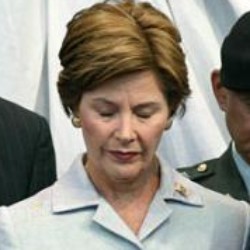} &
  \hspace{-1.0ex}\includegraphics[align=c,width=0.21\columnwidth]{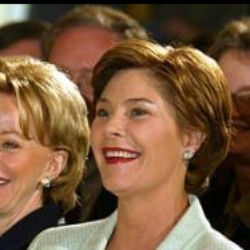} \vspace{3.0pt} \\

  \hspace{-1.0ex}FN &
  \includegraphics[align=c,width=0.21\columnwidth]{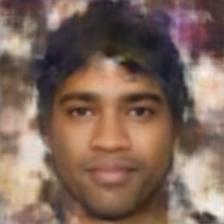} & 
  \hspace{-1.0ex}\includegraphics[align=c,width=0.21\columnwidth]{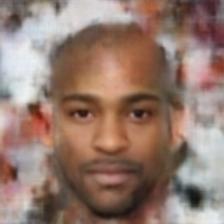} &
  \hspace{-1.0ex}\includegraphics[align=c,width=0.21\columnwidth]{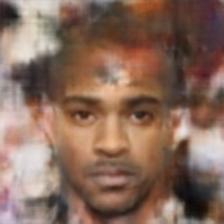} &
  \hspace{-1.0ex}\includegraphics[align=c,width=0.21\columnwidth]{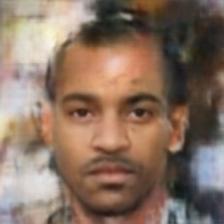} &
  \includegraphics[align=c,width=0.21\columnwidth]{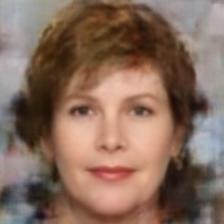} & 
  \hspace{-1.0ex}\includegraphics[align=c,width=0.21\columnwidth]{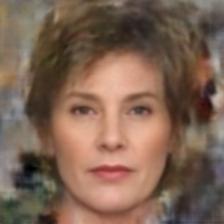} &
  \hspace{-1.0ex}\includegraphics[align=c,width=0.21\columnwidth]{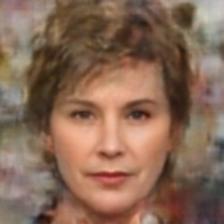} &
  \hspace{-1.0ex}\includegraphics[align=c,width=0.21\columnwidth]{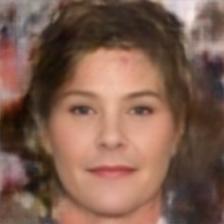} \vspace{3.0pt} \\

  \hspace{-1.0ex}VGG &
  \includegraphics[align=c,width=0.21\columnwidth]{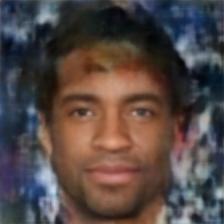} & 
  \hspace{-1.0ex}\includegraphics[align=c,width=0.21\columnwidth]{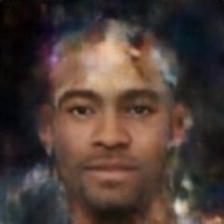} &
  \hspace{-1.0ex}\includegraphics[align=c,width=0.21\columnwidth]{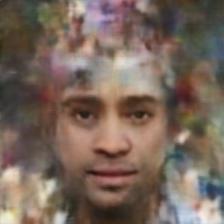} &
  \hspace{-1.0ex}\includegraphics[align=c,width=0.21\columnwidth]{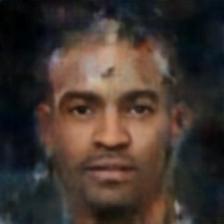} &
  \includegraphics[align=c,width=0.21\columnwidth]{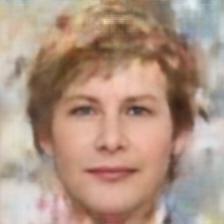} & 
  \hspace{-1.0ex}\includegraphics[align=c,width=0.21\columnwidth]{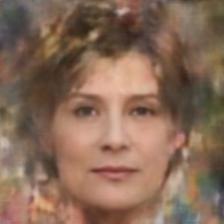} &
  \hspace{-1.0ex}\includegraphics[align=c,width=0.21\columnwidth]{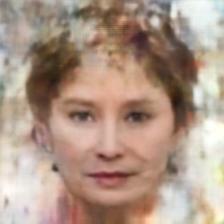} &
  \hspace{-1.0ex}\includegraphics[align=c,width=0.21\columnwidth]{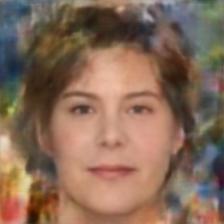} \vspace{3.0pt} \\
  
  \hspace{-1.0ex}\cite{hassner2015effective} &
  \includegraphics[align=c,width=0.11\columnwidth]{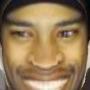} & 
  \hspace{-1.0ex}\includegraphics[align=c,width=0.11\columnwidth]{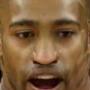} &
  \hspace{-1.0ex}\includegraphics[align=c,width=0.11\columnwidth]{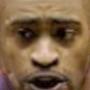} &
  \hspace{-1.0ex}\includegraphics[align=c,width=0.11\columnwidth]{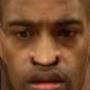} &
  \includegraphics[align=c,width=0.11\columnwidth]{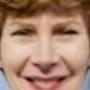} & 
  \hspace{-1.0ex}\includegraphics[align=c,width=0.11\columnwidth]{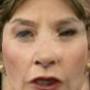} &
  \hspace{-1.0ex}\includegraphics[align=c,width=0.11\columnwidth]{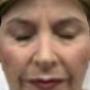} &
  \hspace{-1.0ex}\includegraphics[align=c,width=0.11\columnwidth]{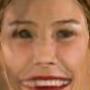} \vspace{3.0pt} \\

\end{tabular}
\caption{Additional face normalization results for the LFW dataset~\cite{huang2007lfw}. Top: input photographs. Middle: result of our method for FaceNet ``avgpool-0'' and VGG-Face ``fc7'' features. Bottom: result of \citet{hassner2015effective}.} \label{fig:frontalization_supplemental}
\end{figure*}

\begin{figure*}
\vspace{-2em}
\begin{tabular}{ccccc|cccc}
  
  \hspace{-1.0ex}Input &
 \includegraphics[align=c,width=0.21\columnwidth]{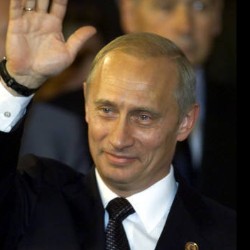} &
  \hspace{-1.0ex}\includegraphics[align=c,width=0.21\columnwidth]{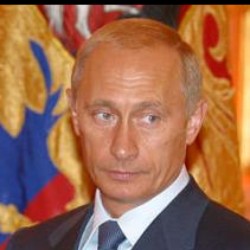} &
  \hspace{-1.0ex}\includegraphics[align=c,width=0.21\columnwidth]{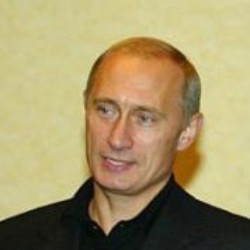} &
  \hspace{-1.0ex}\includegraphics[align=c,width=0.21\columnwidth]{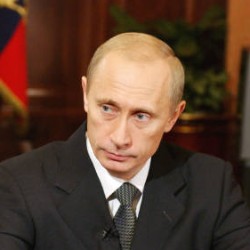} &
  \includegraphics[align=c,width=0.21\columnwidth]{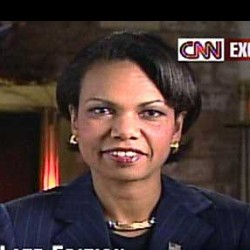} & 
  \hspace{-1.0ex}\includegraphics[align=c,width=0.21\columnwidth]{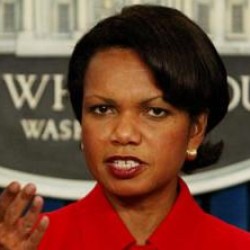} &
  \hspace{-1.0ex}\includegraphics[align=c,width=0.21\columnwidth]{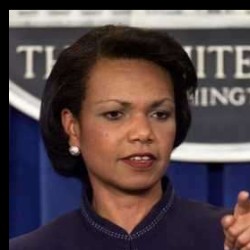} &
  \hspace{-1.0ex}\includegraphics[align=c,width=0.21\columnwidth]{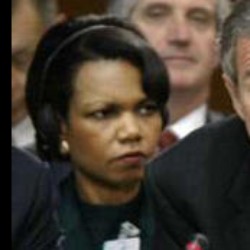} \vspace{3.0pt} \\

  \hspace{-1.0ex}FN &
  \includegraphics[align=c,width=0.21\columnwidth]{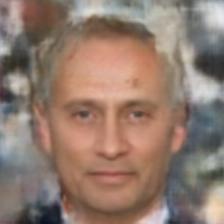} & 
  \hspace{-1.0ex}\includegraphics[align=c,width=0.21\columnwidth]{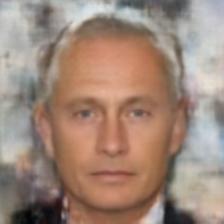} &
  \hspace{-1.0ex}\includegraphics[align=c,width=0.21\columnwidth]{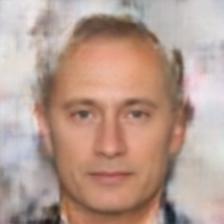} &
  \hspace{-1.0ex}\includegraphics[align=c,width=0.21\columnwidth]{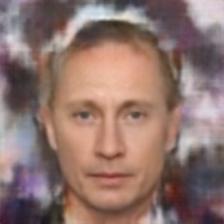} &
  \includegraphics[align=c,width=0.21\columnwidth]{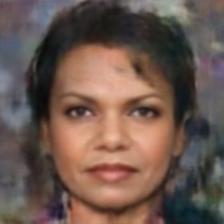} & 
  \hspace{-1.0ex}\includegraphics[align=c,width=0.21\columnwidth]{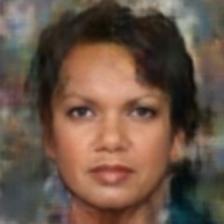} &
  \hspace{-1.0ex}\includegraphics[align=c,width=0.21\columnwidth]{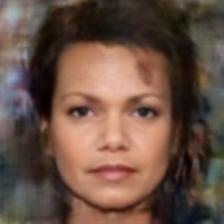} &
  \hspace{-1.0ex}\includegraphics[align=c,width=0.21\columnwidth]{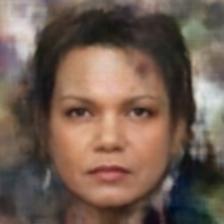} \vspace{3.0pt} \\

  \hspace{-1.0ex}VGG &
  \includegraphics[align=c,width=0.21\columnwidth]{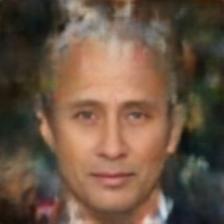} & 
  \hspace{-1.0ex}\includegraphics[align=c,width=0.21\columnwidth]{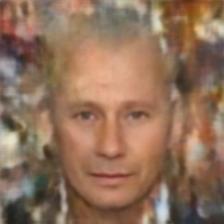} &
  \hspace{-1.0ex}\includegraphics[align=c,width=0.21\columnwidth]{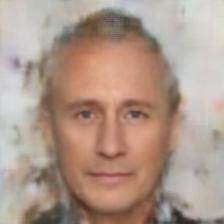} &
  \hspace{-1.0ex}\includegraphics[align=c,width=0.21\columnwidth]{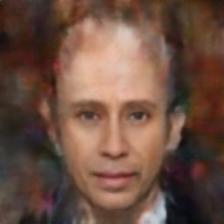} &
  \includegraphics[align=c,width=0.21\columnwidth]{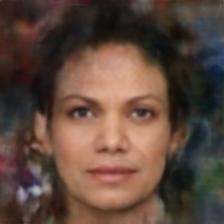} & 
  \hspace{-1.0ex}\includegraphics[align=c,width=0.21\columnwidth]{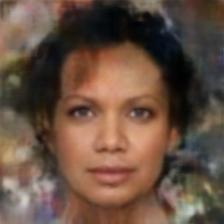} &
  \hspace{-1.0ex}\includegraphics[align=c,width=0.21\columnwidth]{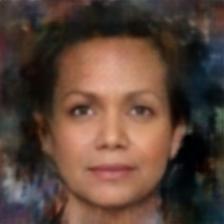} &
  \hspace{-1.0ex}\includegraphics[align=c,width=0.21\columnwidth]{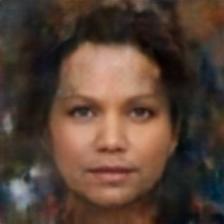} \vspace{3.0pt} \\
  
  \hspace{-1.0ex}\cite{hassner2015effective} &
  \includegraphics[align=c,width=0.11\columnwidth]{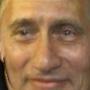} & 
  \hspace{-1.0ex}\includegraphics[align=c,width=0.11\columnwidth]{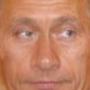} &
  \hspace{-1.0ex}\includegraphics[align=c,width=0.11\columnwidth]{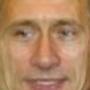} &
  \hspace{-1.0ex}\includegraphics[align=c,width=0.11\columnwidth]{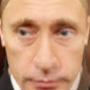} &
  \includegraphics[align=c,width=0.11\columnwidth]{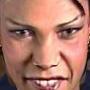} & 
  \hspace{-1.0ex}\includegraphics[align=c,width=0.11\columnwidth]{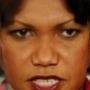} &
  \hspace{-1.0ex}\includegraphics[align=c,width=0.11\columnwidth]{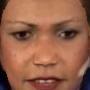} &
  \hspace{-1.0ex}\includegraphics[align=c,width=0.11\columnwidth]{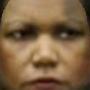} \vspace{3.0pt} \\
  
  \hline
  
  \hspace{-1.0ex}Input &
 \includegraphics[align=c,width=0.21\columnwidth]{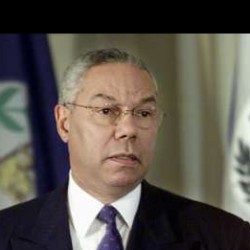} &
  \hspace{-1.0ex}\includegraphics[align=c,width=0.21\columnwidth]{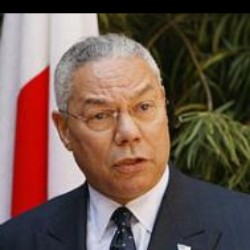} &
  \hspace{-1.0ex}\includegraphics[align=c,width=0.21\columnwidth]{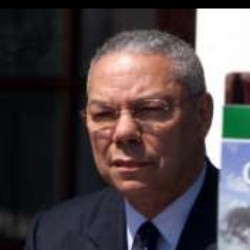} &
  \hspace{-1.0ex}\includegraphics[align=c,width=0.21\columnwidth]{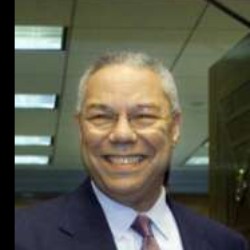} &
  \includegraphics[align=c,width=0.21\columnwidth]{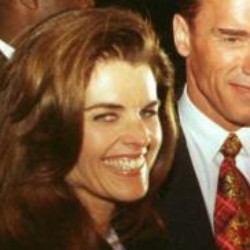} & 
  \hspace{-1.0ex}\includegraphics[align=c,width=0.21\columnwidth]{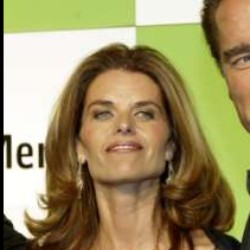} &
  \hspace{-1.0ex}\includegraphics[align=c,width=0.21\columnwidth]{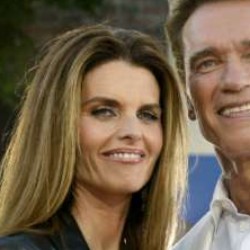} &
  \hspace{-1.0ex}\includegraphics[align=c,width=0.21\columnwidth]{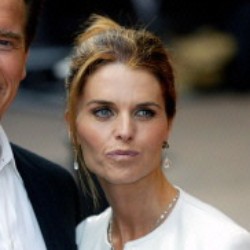} \vspace{3.0pt} \\

  \hspace{-1.0ex}FN &
  \includegraphics[align=c,width=0.21\columnwidth]{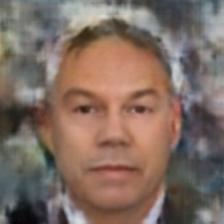} & 
  \hspace{-1.0ex}\includegraphics[align=c,width=0.21\columnwidth]{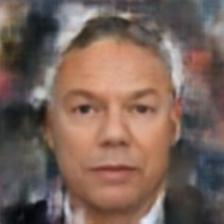} &
  \hspace{-1.0ex}\includegraphics[align=c,width=0.21\columnwidth]{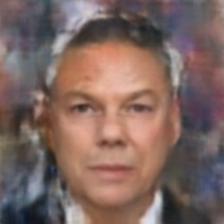} &
  \hspace{-1.0ex}\includegraphics[align=c,width=0.21\columnwidth]{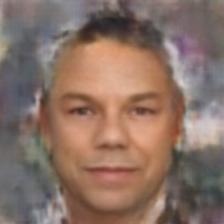} &
  \includegraphics[align=c,width=0.21\columnwidth]{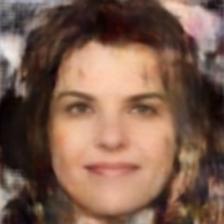} & 
  \hspace{-1.0ex}\includegraphics[align=c,width=0.21\columnwidth]{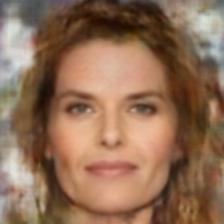} &
  \hspace{-1.0ex}\includegraphics[align=c,width=0.21\columnwidth]{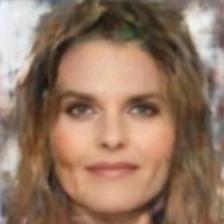} &
  \hspace{-1.0ex}\includegraphics[align=c,width=0.21\columnwidth]{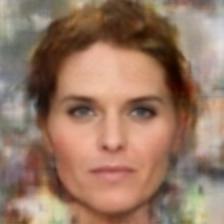} \vspace{3.0pt} \\

  \hspace{-1.0ex}VGG &
  \includegraphics[align=c,width=0.21\columnwidth]{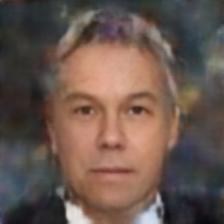} & 
  \hspace{-1.0ex}\includegraphics[align=c,width=0.21\columnwidth]{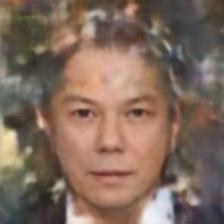} &
  \hspace{-1.0ex}\includegraphics[align=c,width=0.21\columnwidth]{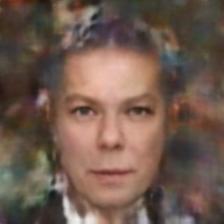} &
  \hspace{-1.0ex}\includegraphics[align=c,width=0.21\columnwidth]{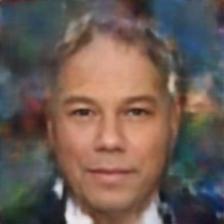} &
  \includegraphics[align=c,width=0.21\columnwidth]{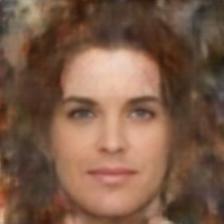} & 
  \hspace{-1.0ex}\includegraphics[align=c,width=0.21\columnwidth]{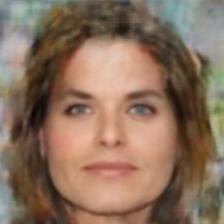} &
  \hspace{-1.0ex}\includegraphics[align=c,width=0.21\columnwidth]{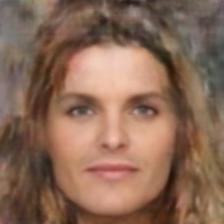} &
  \hspace{-1.0ex}\includegraphics[align=c,width=0.21\columnwidth]{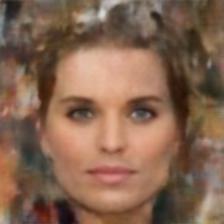} \vspace{3.0pt} \\
  
  \hspace{-1.0ex}\cite{hassner2015effective} &
  \includegraphics[align=c,width=0.11\columnwidth]{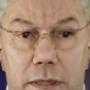} & 
  \hspace{-1.0ex}\includegraphics[align=c,width=0.11\columnwidth]{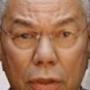} &
  \hspace{-1.0ex}\includegraphics[align=c,width=0.11\columnwidth]{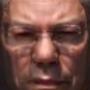} &
  \hspace{-1.0ex}\includegraphics[align=c,width=0.11\columnwidth]{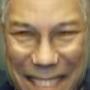} &
  \includegraphics[align=c,width=0.11\columnwidth]{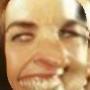} & 
  \hspace{-1.0ex}\includegraphics[align=c,width=0.11\columnwidth]{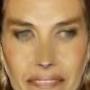} &
  \hspace{-1.0ex}\includegraphics[align=c,width=0.11\columnwidth]{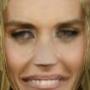} &
  \hspace{-1.0ex}\includegraphics[align=c,width=0.11\columnwidth]{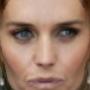} \vspace{3.0pt} \\
  
\end{tabular}
\caption{Additional face normalization results similar to Fig.~\ref{fig:frontalization_supplemental}} \label{fig:frontalization2}
\end{figure*}

\begin{figure*}
\begin{tabular}{cccc}
  \includegraphics[width=0.22\textwidth]{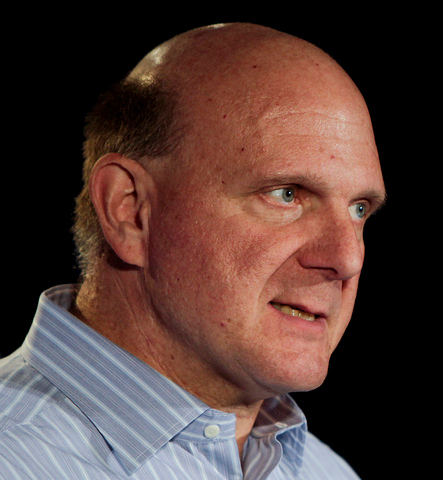} &
  \includegraphics[width=0.22\textwidth]{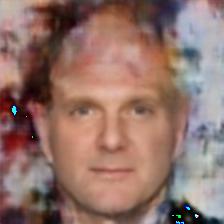} &
  \includegraphics[width=0.22\textwidth]{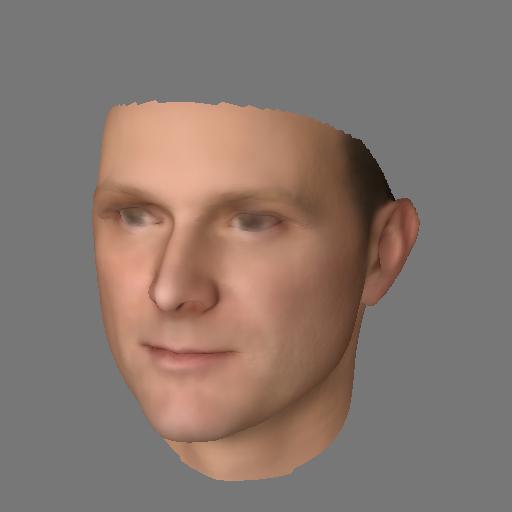} &
  \includegraphics[width=0.22\textwidth]{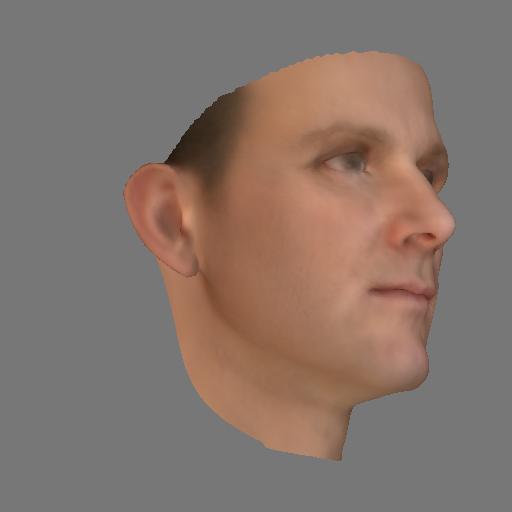} \\
  \includegraphics[width=0.22\textwidth]{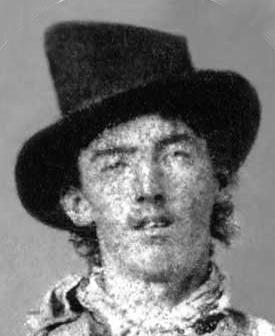} &
  \includegraphics[width=0.22\textwidth]{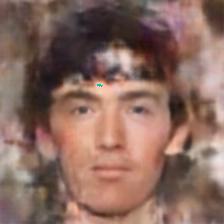} &
  \includegraphics[width=0.22\textwidth]{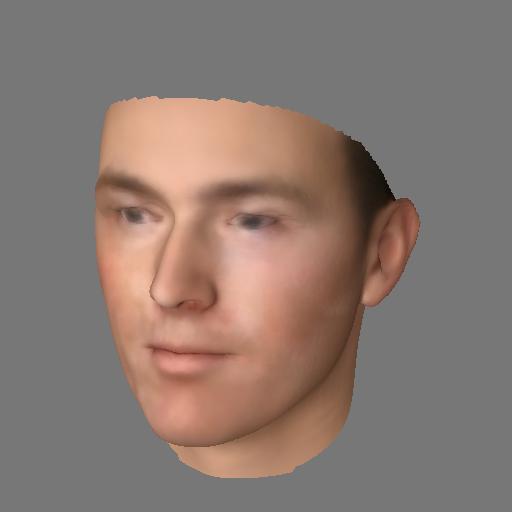} &
  \includegraphics[width=0.22\textwidth]{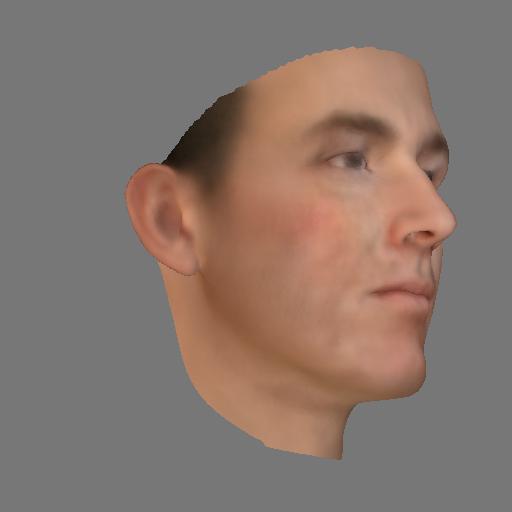} \\
  \includegraphics[width=0.22\textwidth]{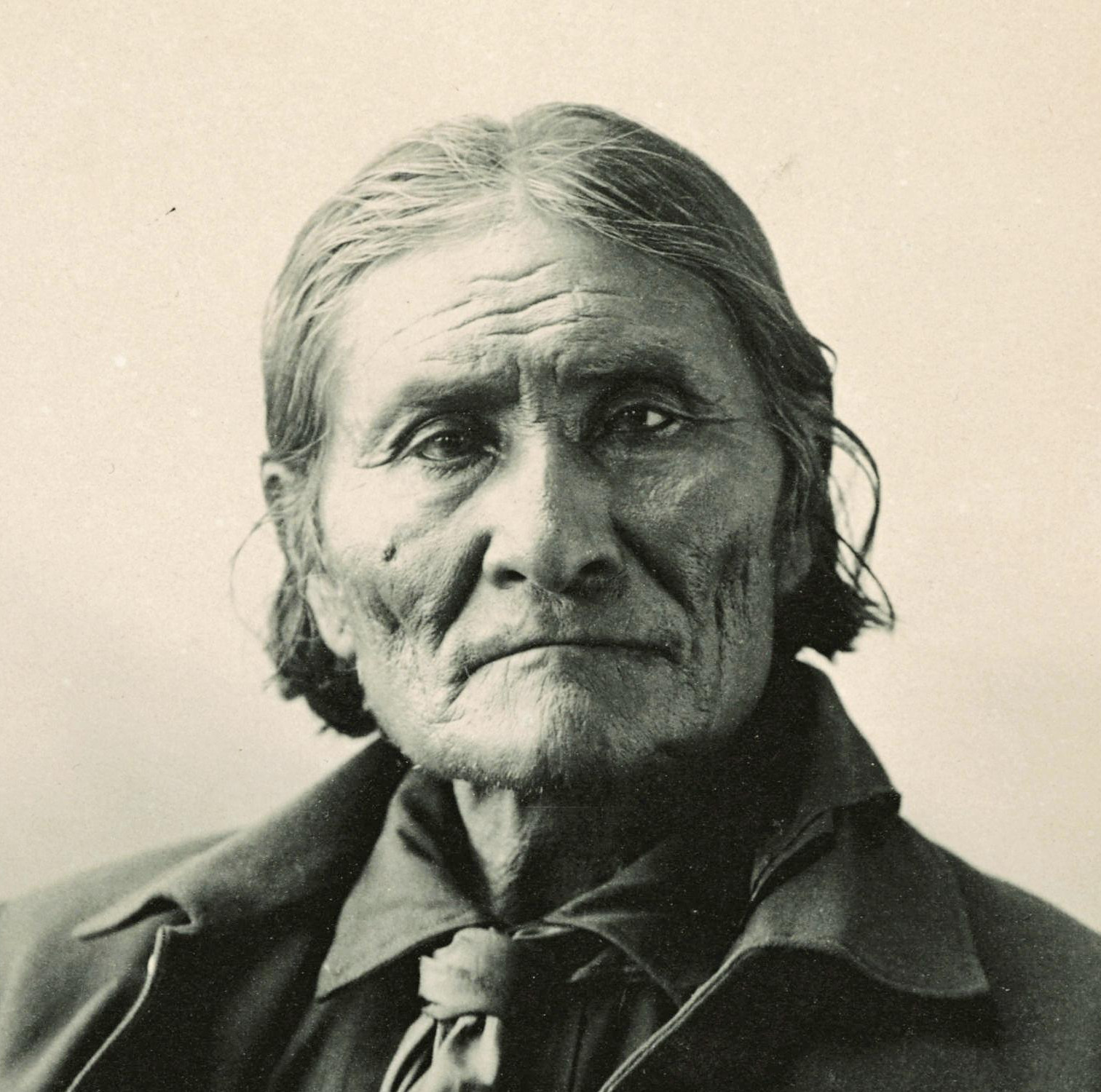} &
  \includegraphics[width=0.22\textwidth]{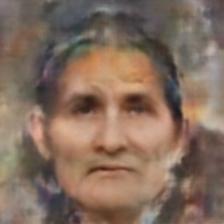} &
  \includegraphics[width=0.22\textwidth]{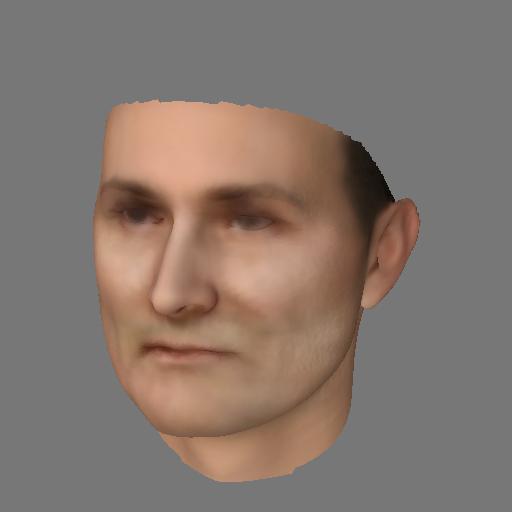} &
  \includegraphics[width=0.22\textwidth]{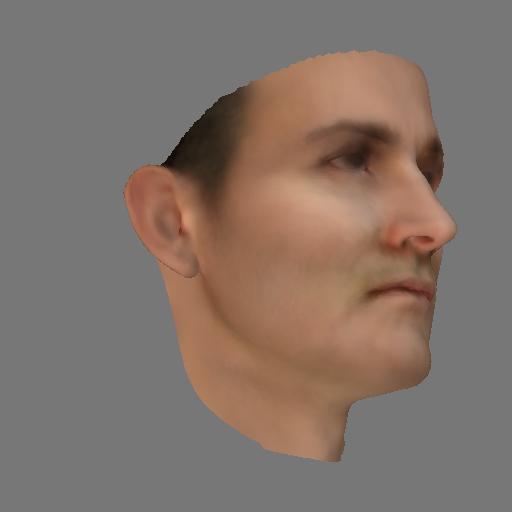} \\

  \includegraphics[width=0.22\textwidth]{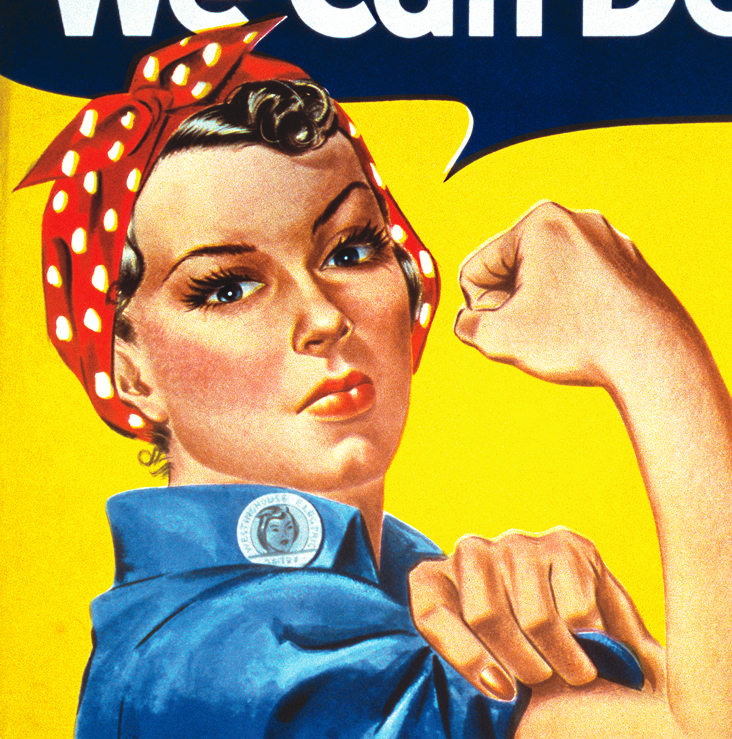} &
  \includegraphics[width=0.22\textwidth]{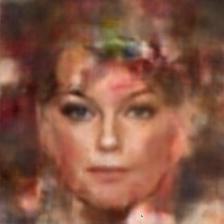} &
  \includegraphics[width=0.22\textwidth]{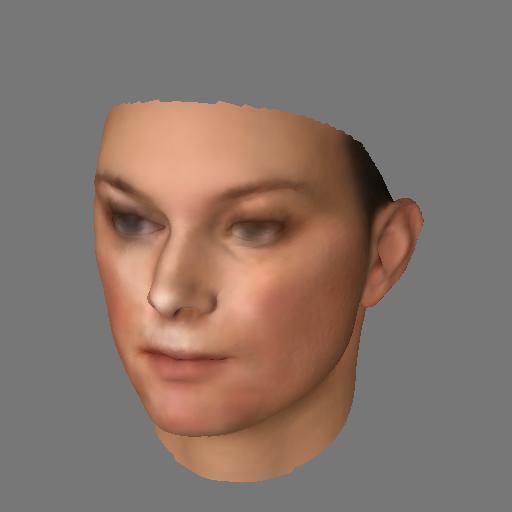} &
  \includegraphics[width=0.22\textwidth]{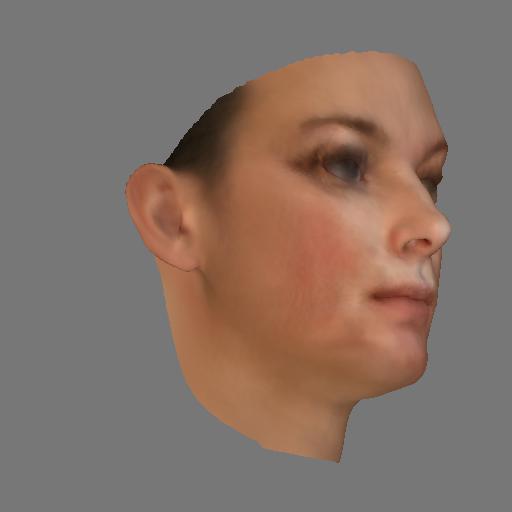} \\
  
\end{tabular}
\caption{Though the model was only trained on natural images, it is robust enough to be applied to degraded photographs and illustrations. 
 Column 1: input image. Column 2: generated 2-D image. Columns 3 and 4: images of 3-D reconstruction taken from 2 different angles.}
\label{fig:3dresults_supplemental}
\end{figure*}

The correspondence could be determined completely manually, but we choose to find the it automatically by rendering the mean face and extracting landmarks from the rendered image (Fig~\ref{fig:mean-face-landmarks}). 

The corresponding vertex for each landmark is found by measuring screen-space distance between the computed landmarks and the projected vertices. This projection is noisy around grazing angles and may pick back-facing vertices or other poor choices. To make the correspondence cleaner, we compute the correspondences separately for multiple, randomly jittered camera matrices, then use voting to determine the most stable matching vertex for each landmark. The final result is a set of 65 vertex indices.

\subsection{Shape Fitting}

Given a set of $65\times2$ matrix of landmark points $L$, our goal is to optimize for the best matching set of 199 shape coefficients $\mathbf{s}$. To find $\mathbf{s}$, we imagine that the landmarks $L$ are the projection of their corresponding vertices $V$, where the $65\times2$ matrix $V$ is defined by the shape parameters $\mathbf{s}$, a translation vector $\mathbf{t}$, a uniform scaling factor $\sigma$, and a fixed projection matrix $P$, as follows. 

Let the $65\times3$ matrix of object-space vertex positions $V_w$ be:

\begin{equation}
    V_w = \begin{bmatrix} B^x \mathbf{s} & B^y \mathbf{s} & B^z \mathbf{s}\end{bmatrix} + \mu
\end{equation}

where $B^{x,y,z}$ are the $65\times199$ morphable model basis matrices and $\mu$ is the $65\times3$ matrix of mean vertex positions.

The $4\times4$ projection matrix $P$ is a perspective projection with a field of view of $10^\circ$ to roughly match the perspective of the training images. The $4\times4$ modelview matrix $M$ is defined by the translation $\mathbf{t}$ and scaling $\sigma$ as:

\begin{equation}
    M = \begin{bmatrix} \sigma & 0 & 0 & t^x \\
    0 & \sigma & 0 & t^y \\
    0 & 0 & \sigma & t^z \\
    0 & 0 & 0 & 1
    \end{bmatrix}
\end{equation}

Given $P$ and $M$, the $65\times4$ matrix of post-projection vertices $V_p$ is defined as:

\begin{equation}
    V_p = \begin{bmatrix}V_w & \mathbf{1}\end{bmatrix} M^T P^T
\end{equation}

and the final, $65\times2$ vertex position matrix $V$ is found by perspective division:

\begin{equation}
    V = \begin{bmatrix}\frac{\mathbf{x}_p}{\mathbf{w}_p} & \frac{\mathbf{y}_p}{\mathbf{w}_p}\end{bmatrix}
\end{equation}

where $\mathbf{x}_p$, $\mathbf{y}_p$, and $\mathbf{w}_p$ are first, second, and fourth columns of $V_p$.

Finally, we optimize for $\mathbf{s}$ using gradient descent with the loss function:

\begin{equation}
    f(\mathbf{s}) = \lVert L - V \rVert^2 + \lambda \lVert \mathbf{s} \rVert^2
\end{equation}

where length term for $\mathbf{s}$ regularizes the optimization towards the mean face (i.e., $\mathbf{s} = \mathbf{0}$), and $\lambda = 0.001$ in our experiments.

\subsection{Fitting Texture}

Once the shape parameters and pose of the model are found, we project the remaining $\approx 53$K vertices of the mesh onto the synthesized face image. The projection produces a $53K\times3$ matrix of vertex colors $C_p$. 

Due to noise in the synthesized image and the inherent inaccuracy of projection at grazing angles, the colors $C_p$ have ugly artifacts. To repair the artifacts, we compute a confidence value $\alpha_i$ at each vertex that downweights vertices outside the facial landmarks and vertices at grazing angles:
\begin{equation}
    \alpha_i = m(x_i, y_i) (1.0 - n_i^z)
\end{equation}

where $m$ is a mask image that is $1$ inside the convex hull of the landmark points and smoothly decays to $0$ outside, and $n_i^z$ is the $z$ component of the $i^{th}$ vertex normal.

Using the confidences, we project the vertex colors $C_p$ onto the morphable model color basis. Let $\mathbf{c}_p$ be the 160K vector produced by flattening $C_p$, $\mathbf{a}$ be the 160K vector produced by repeating the confidences $\alpha_i$ for each color channel, and $A$ be the $160K\times199$ matrix of confidences produced by tiling $\mathbf{a}$. The 199 color parameters $\mathbf{z}$ are found by solving an over-constrained linear system in the least-squares sense:

\begin{equation}
    \begin{bmatrix}(B \circ A) \\
    \lambda I\end{bmatrix} \mathbf{z} = \begin{bmatrix}(\mathbf{c_p} - \mu) \circ \mathbf{a} \\
    \mathbf{0}\end{bmatrix}
\end{equation}

where $\circ$ represents the element-wise product, $B$ is the $160K\times199$ color basis matrix, $I$ is the identity matrix, $\mu$ is the model's mean color vector, and $\lambda$ is a regularization constant. 

The flattened model color vector $\mathbf{c}_b$ is found by un-projecting $\mathbf{z}$:
\begin{equation}
    \mathbf{c}_b = B^T \mathbf{z} + \mu
\end{equation}

and the final flattened color vector $\mathbf{c}$ is defined by interpolating between the projected and model colors:

\begin{equation}
    \mathbf{c} = \mathbf{c}_p \circ \mathbf{a} + \mathbf{c}_b \circ (\mathbf{1} - \mathbf{a})
\end{equation}

\section{Automatic Photo Adjustment}
Let $\mathbf{m}_P$ and $\mathbf{m}_N$ be the mean face colors for the input and normalized images, respectively. Our adjusted image is computed using a per-channel, piecewise-linear color shift function $r^c(\mathbf{p})$ over the pixels of $P$:
\begin{equation}
r^c(\mathbf{p}) = \left\{ \begin{array}{rcl} 
\mathbf{p}^c \frac{\mathbf{m}^c_N}{\mathbf{m}^c_P} & \mbox{if} & \mathbf{p}^c <= \mathbf{m}^c_P \\
1 - (1 - \mathbf{p}^c) \frac{1 - \mathbf{m}^c_N}{1 - \mathbf{m}^c_P} & \mbox{if} & \mathbf{p}^c > \mathbf{m}^c_P, \\
\end{array} \right \}
\end{equation}
where $c$ are the color channels. We chose YCrCb as the color representation in our experiments.

\end{document}